\documentclass[preprint,3p,times,twocolumn]{elsarticle}

\usepackage{fancyhdr}
\usepackage{url}

\fancypagestyle{pprintTitle}{%
  \fancyhead[L]{\small Published in Engineering Applications of Artificial Intelligence \space}
  \fancyhead[R]{\small DOI: \url{https://doi.org/10.1016/j.engappai.2024.109030}}
  \fancyfoot[C]{}
}

\usepackage{physics}
\usepackage{multirow}
\usepackage{graphicx}
\usepackage{hyperref}
\usepackage{xcolor}

\hypersetup{
    colorlinks=true,
    linkcolor=red,
    filecolor=magenta,      
    urlcolor=cyan,
    citecolor=green,
}

\usepackage{listings}
\usepackage[normalem]{ulem}
\usepackage{caption}

\usepackage{algorithm,algorithmic}
\usepackage{relsize}
\usepackage{balance}

\usepackage[flushleft]{threeparttable}

\usepackage{amsmath}
\usepackage{amssymb}

\usepackage{subcaption}

\newcommand\ie{\textit{i.e.}}
\newcommand\eg{\textit{e.g.}}

\usepackage[acronym, shortcuts]{glossaries}

\definecolor{codegreen}{rgb}{0,0.6,0}
\definecolor{codegray}{rgb}{0.5,0.5,0.5}
\definecolor{codepurple}{rgb}{0.58,0,0.82}
\definecolor{backcolour}{rgb}{0.95,0.95,0.92}

\lstdefinestyle{mystyle}{
    backgroundcolor=\color{backcolour},   
    commentstyle=\color{codegreen},
    keywordstyle=\color{magenta},
    numberstyle=\tiny\color{codegray},
    stringstyle=\color{codepurple},
    basicstyle=\ttfamily\footnotesize,
    breakatwhitespace=false,         
    breaklines=true,                 
    captionpos=b,                    
    keepspaces=true,                 
    numbers=left,                    
    numbersep=5pt,                  
    showspaces=false,                
    showstringspaces=false,
    showtabs=false,                  
    tabsize=2
}

\makeglossaries
\newacronym{nas}{NAS}{Neural Architecture Search}
\newacronym{fcn}{FCN}{Fully Convolutional Network}
\newacronym{kd}{KD}{Knowledge Distillation}
\newacronym{pkd}{PKD}{Pseudoknowledge Distillation}
\newacronym{cc}{CC}{Pearson’s Correlation Coefficient}
\newacronym{kld}{KLD}{Kullback-Leibler Divergence}
\newacronym{nss}{NSS}{Normalized Scanpath Saliency}
\newacronym{sim}{SIM}{Similarity}
\newacronym{auc}{AUC}{Area under ROC Curve}

\journal{Engineering Applications of Artificial Intelligence}

\begin{document}

\begin{frontmatter}

\title{SalNAS: Efficient Saliency-prediction Neural Architecture Search with self-knowledge distillation}

\author[a]{Chakkrit Termritthikun\corref{contrib-6ee20f3bbdae22c3f602de1fbc2002b8}}
\ead{chakkritt@nu.ac.th}\cortext[contrib-6ee20f3bbdae22c3f602de1fbc2002b8]{Corresponding author.}

\author[b]{Ayaz~Umer}
\ead{ayaz.umer@mymail.unisa.edu.au}

\author[c]{Suwichaya Suwanwimolkul}
\ead{suwichaya.s@chula.ac.th}

\author[d]{Feng Xia}
\ead{f.xia@ieee.org}

\author[b]{Ivan Lee}
\ead{ivan.lee@unisa.edu.au}

 \address[a]{ School of Renewable Energy and Smart Grid Technology (SGtech) \unskip, Naresuan University\unskip, Phitsanulok\unskip, 65000\unskip, Thailand}

 \address[b]{STEM\unskip, 
    University of South Australia\unskip, Adelaide \unskip, SA \unskip, 5095 \unskip, Australia}

\address[c]{Department of Electrical Engineering\unskip, 
    Faculty of Engineering\unskip, Chulalongkorn University\unskip, Bangkok\unskip, 10330\unskip, Thailand}
    
 \address[d]{School of Computing Technologies\unskip, 
    RMIT University\unskip, Melbourne \unskip, VIC \unskip, 3000 \unskip, Australia}

\begin{abstract}
Recent advancements in deep convolutional neural networks have significantly improved the performance of saliency prediction. However, the manual configuration of the neural network architectures requires domain knowledge expertise and can still be time-consuming and error-prone. To solve this, we propose a new \acrfull{nas} framework for saliency prediction with two contributions. Firstly, a supernet for saliency prediction is built with a weight-sharing network containing all candidate architectures, by integrating a dynamic convolution into the encoder-decoder in the supernet, termed SalNAS. Secondly, despite the fact that SalNAS is highly efficient (20.98 million parameters), it can suffer from the lack of generalization. To solve this, we propose a self-knowledge distillation approach, termed Self-KD, that trains the student SalNAS with the weighted average information between the ground truth and the prediction from the teacher model. The teacher model, while sharing the same architecture, contains the best-performing weights chosen by cross-validation. Self-KD can generalize well without the need to compute the gradient in the teacher model, enabling an efficient training system. By utilizing Self-KD, SalNAS outperforms other state-of-the-art saliency prediction models in most evaluation rubrics across seven benchmark datasets while being a lightweight model. The code will be available at \href{https://github.com/chakkritte/SalNAS}{https://github.com/chakkritte/SalNAS}.
\end{abstract}


\end{frontmatter}



\pagestyle{fancy}
\fancyhf{}
\fancyhead[L]{Published in Engineering Applications of AI}
\fancyhead[R]{DOI: \url{https://doi.org/10.1016/j.engappai.2024.109030}}

\section{Introduction}

Deep learning models have achieved state-of-the-art performance in several domains, including saliency prediction. The goal of saliency prediction is to provide visual attention that highlights certain areas that are visually meaningful in an image.  Visual attention or saliency prediction represents an important mechanism of the human visual system, enabling the interpretation of the most relevant information in the visual scene. The procedure of the saliency prediction is as follows: Given an input image, the prediction produces a saliency map, indicating its important regions by marking it with high intensity. Thus, the resulting visual attention plays an important role in many real-world applications, including object detection  \cite{cho2021human,tu2023multimodal, zhu2024learning, liu2023novel, liu2023part}, segmentation \cite{li2019constrained, li2020personal}, driver focus of attention \cite{huang2022prediction}, and radiologist gaze prediction \cite{lou2023predicting}. Early research on saliency prediction mainly relies on hand-crafted features. The limitation of these hand-crafted features is their ineffectiveness towards complex scenes, \ie, natural images with multiple objects or multiple salient regions \cite{hong2022high}.     

Recently, deep learning has been used for its powerful feature representation. To overcome the limitation in complex scenes, the recent works extract the hierarchical features that can predict the saliency map in a multi-scale fashion~\cite{kroner2020contextual, jia2020eml, wang2021saled, umer2022device, hu2021fastsal,liu2018learningto, he2020exploring}.   The hierarchical feature is extracted using a \acrfull{fcn} \cite{long2015fully} where the convolutional layers are stacked into a hierarchical structure. The model architectures consist of an encoder-decoder framework where the encoder is used as the backbone, and the entire framework can be trained in an end-to-end fashion.
However, the architectures of these decoders are designed manually, which could provide high accuracy but are not well-optimized for other properties, especially for latency and power consumption.

To address such problems, neural architecture search \cite{zoph2018learning}   can be used in searching for the best architectures for the best fitting towards the ground truth and the computational resources. One can argue that most of the architectures of the (pre-trained) encoders were optimized with \acrshort{nas}, at one point. However, the architectures are optimized for classification tasks \cite{cai2019once,wang2021attentivenas,tan2019efficientnet, shi2023ebnas}. Therefore, their performance is not tailored for saliency prediction. In addition, optimizing only the encoder with \acrshort{nas} will require the post-training where, for every optimal choice of the encoder, each pair of encoder and decoder will need to be trained again for saliency prediction. 

In this paper, we propose to build a framework where the encoder and decoder are optimized together with \acrshort{nas}, termed SalNAS. Our framework not only allows both the encoder and decoder to be trained jointly,  but also optimized for various properties, \eg,  accuracy, number of parameters, and computational complexity. We build a supernet that employs dynamic convolution in both the encoder and decoder, which allows the customization of the kernel dimensions for each candidate pair.

The supernet assembles all the candidate architectures as subnets whose weights are shared as inspired by~\cite{cai2019once}, resulting in a weight-sharing \acrshort{nas}.  Thus, the best architecture suitable for different computing platforms can be obtained by a search algorithm. Our resulting SalNAS architecture is highly efficient, but the small neural network still struggles to generalize effectively across diverse datasets due to the limited number of parameters. Therefore, we propose a \acrfull{kd} method called Self-KD that regularizes the knowledge from the student SalNAS with the weighted average information between the ground truth and the teacher model.  The teacher model is a neural network whose weights are the average of the best-performing students obtained from different epochs. The best-performing students are chosen via cross-validation. In addition to  \acrshort{kd}, our learning loss is derived from saliency evaluation metrics, whose values are in different ranges~\cite{bylinskii2018different}. Thus, we also propose to adjust the value of the loss functions to the same range $(0,1)$, which results in a new learning loss that improves the overall learning capabilities. Together with Self-KD, our model SalNAS offers the best architecture transfer and achieves the highest performance. Therefore, our contributions are as follows:  
 
 \vspace{1em}

\begin{itemize}\setlength\itemsep{1em}
\item We proposed a \acrshort{nas}-based method to build a supernet for saliency prediction, termed SalNAS. Our method uses dynamic convolutional layers for the encoder and decoder to build a supernet. This allows multiple encoder-decoder pairs to be trained only once with a weight-sharing strategy. The best architecture suitable for different computing platforms can be obtained by any search algorithm.  
 
\item We proposed Self-KD to enhance the generalization by regularizing the knowledge from the model itself. That is, the student's knowledge is regularized by one of the best performance models chosen via cross-validation. This leads to performance gain across most evaluation metrics and different backbone sizes---from small to large.

\item Our proposed method outperforms the state-of-the-art saliency prediction models, across \textit{five performance evaluation metrics} --- \acrfull{cc}, \acrfull{kld}, \acrfull{nss}, \acrfull{sim}, and \acrfull{auc} --- and  \textit{ seven computational consumption measurements} --- computational complexity, parameters, model size, carbon emission, power consumption, latency, and throughput.
\end{itemize}

\section{Related Work}
Saliency prediction can typically be divided into two categories based on their simulated attention processes: bottom-up (free-viewing) and top-down (task-driven). Our work aims to anticipate computationally efficient saliency prediction on raw images to simulate free-viewing \textit{bottom-up} visual attention. Thus, we mainly focus on the related works for free-viewing computationally efficient saliency prediction models.

\subsection{Compact  Saliency Prediction.}  Early research in saliency prediction relies on manually extracting high-dimensional features to identify important areas that represent attention. A work \cite{liang2015predicting} utilized a combination of multi-scale low-level features, including color, intensity, and orientation, to compute a saliency map. The work \cite{harel2006graph} computes a saliency map using a graph-based approach using Markov chains. Other examples of these works are \cite{borji2012boosting,zhang2013saliency}. However, they fail to produce saliency prediction for complex scenes. Deep learning~\cite{vig2014large}  has been shown to overcome complex scenes by training with large-scale annotated data.  The architecture contains a pair of encoders and decoders~\cite{vig2014large, kroner2020contextual, jia2020eml, wang2021saled, umer2022device, hu2021fastsal}. Given input images, the encoders perform the latent feature extraction; meanwhile, the decoder converts these features into visual attention, \eg,  MSI \cite{kroner2020contextual}, EML-Net \cite{jia2020eml}, SalED \cite{wang2021saled}, \acrshort{pkd} \cite{umer2022device}, and FastSal \cite{hu2021fastsal}. To maximize the efficiency, the following pre-trained models can be used as the encoder, \ie, EfficientNet \cite{tan2019efficientnet}, OFA-595 \cite{cai2019once}, and ResNet \cite{he2016deep}.    FastSal \cite{hu2021fastsal} utilizes a lightweight backbone, \ie, MobileNetV2 \cite{sandler2018mobilenetv2}, and EfficientNet-B0 \cite{tan2019efficientnet}, to achieve the fast saliency prediction offering low latency for lightweight computing devices. Another work \cite{zabihi2022compact} proposed a real-time saliency prediction by introducing a modified U-Net architecture and location-dependent fully connected layers to build a fast saliency model suitable for edge devices. \acrshort{pkd} \cite{umer2022device} proposed to distill knowledge from a bigger network (teacher) to a smaller (student) network with higher efficiency.

\subsection{Neural Architecture Search}
\textit{Conventional Neural Architectures}  find candidate architectures by employing \textit{(i) a supernet}, a large-size network containing millions of subnets, and \textit{(ii) an architecture searching}, a  technique for finding candidate subnets, either by reinforcement learning~\cite{baker2017RLNAS, zoph2017RLNAS} or evolutionary search~\cite{sun2020evolving, termritthikun2023evolutionary}. Yet, these techniques require high latency. To reduce the latency, recent works~\cite{liu2018DART, xu2020DART} proposed gradient-based optimization to search for subnets in a supernet which is a \textit{directed acyclic graph}, termed DARTS. Recently, Saliency-aware \acrshort{nas}~\cite{saliencyaware2022NIPS}  proposed to employ the saliency information to further improve the training in a gradient-based \acrshort{nas} similar to DARTS.  The architecture and weights of the directed acyclic graph are optimized in association with the intermediate results of the saliency map.  Nevertheless, the directed acyclic graph is extremely long ---each layer contains multiple configurations, \eg, candidate operations and parameters, which results in high memory and latency issues in training.

\textit{High-efficiency approach.}  To solve the memory and latency problem in the conventional \acrshort{nas}, the high-efficiency approaches~\cite{brockICLR2018oneshot,benderPMLR2018oneshot, cai2019once, yu2020bignas} employ a supernet, a weight-sharing sequential network that replaces the directed acyclic graph. One-shot approaches~\cite{brockICLR2018oneshot,benderPMLR2018oneshot} use a \textit{sequential network} as a new supernet, whose parameters between subnets are shared. Once-for-All~\cite{cai2019once} further improves the efficiency by proposing a progressive shrinking that prunes the supernet for the candidate subnets compatible with various hardware platforms. However,  additional tuning is required. This problem is solved by BigNAS~\cite{yu2020bignas} that concurrently trained all subnets in a single-stage model, allowing high-quality subnets to be sliced without post-tuning. Uniform sampling is generally used for its simplicity, but affecting the final accuracy. This issue is solved by an attentive sampling~\cite{wang2021attentivenas}. 

\subsection{Knowledge Distillation (KD)}
  
\textit{Knowledge distillation} focuses on \textit{transferring} the knowledge from the more complex and bigger teacher models to improve the generalization performance of the students. It has shown to be beneficial in many applications, \eg, ~\cite{umer2022device},~\cite{hu2021fastsal},~\cite{termritthikun2023explainable}. \acrshort{kd} can be used in \acrshort{nas}, which automates the design of neural architecture. For example, the work~\cite{kang2020towards} proposed a novel framework of \acrshort{kd} and \acrshort{nas} by introducing \acrshort{kd} loss to facilitate model search and distillation by using an ensemble-based teacher network. Another work ~\cite{li2020block} utilizes \acrshort{nas}, which divides the search space into blocks for efficient training and utilizes a novel distillation approach to supervise architecture search in a block-wise pattern. 


\begin{figure*}[t!]
\begin{center}
\includegraphics[width=1\linewidth]{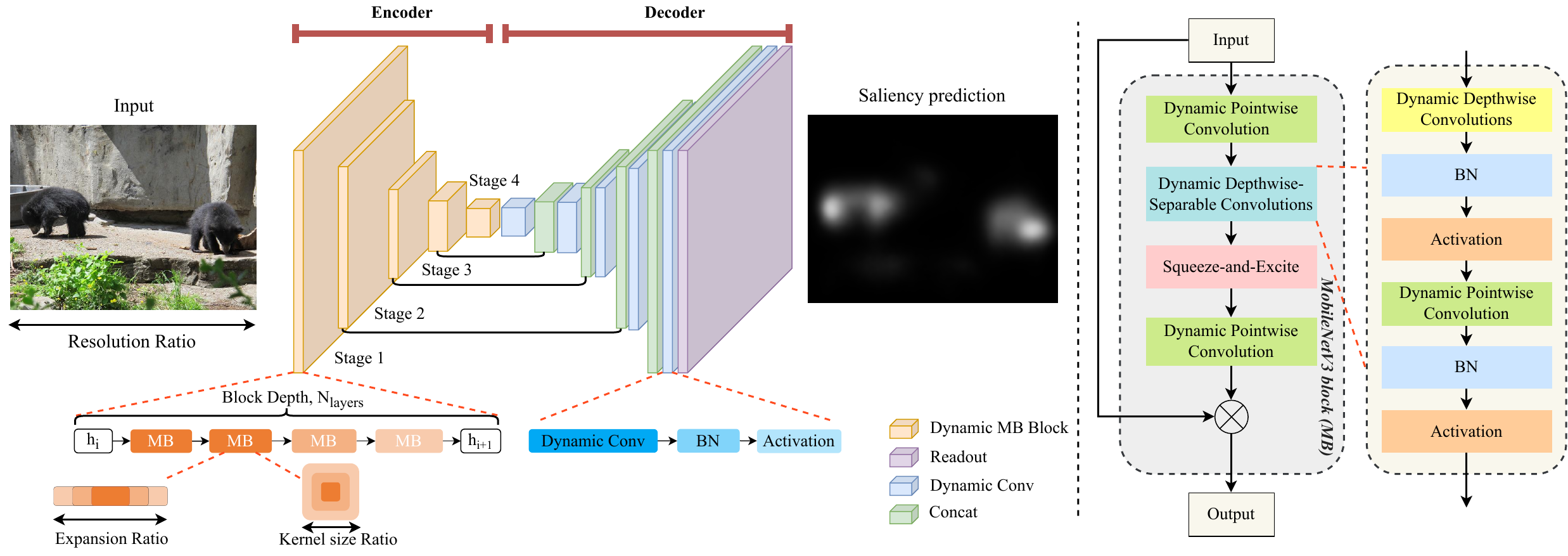}
\end{center}
\caption{Our proposed architecture consists of the encoder and decoder modules that employ dynamic convolutional layers. }
\label{fig:1}
\end{figure*}

\begin{table*}[t]
\centering
\caption{Supernet Search Space: Configuration Illustration and Design Choices for Architecture Search}
\label{tab:1}
\resizebox{\textwidth}{!}{%
\begin{tabular}{lcccccc}
\hline
\textbf{Block} & \textbf{Width} & \textbf{Depth} & \textbf{Kernel size} & \textbf{Expansion} & \textbf{Squeeze and Excitation} & \textbf{Stride} \\ \hline
First Convolution      & \{16,24\}           & -               & 3       & -         & - & 2 \\
MBConv-1         & \{16,24\}           & \{1,2\}         & \{3,5\} & 1         & N & 1 \\
MBConv-2         & \{24,32\}           & \{3,4,5\}       & \{3,5\} & \{4,5,6\} & N & 2 \\
MBConv-3         & \{32,40\}           & \{3,4,5,6\}     & \{3,5\} & \{4,5,6\} & Y & 2 \\
MBConv-4         & \{62,70\}           & \{3,4,5,6\}     & \{3,5\} & \{4,5,6\} & N & 2 \\
MBConv-5         & \{112,120,128\}     & \{3,4,5,6,7,8\} & \{3,5\} & \{4,5,6\} & Y & 1 \\
MBConv-6         & \{192,200,208,216\} & \{3,4,5,6,7,8\} & \{3,5\} & 6         & Y & 2 \\
MBConv-7         & \{216,224\}         & \{1,2\}         & \{3,5\} & 6         & Y & 1 \\ \hline
Last Convolution      & \{1296,1344\}       & -               & 1       & 6         & - & 1 \\ \hline
Input resolution & \multicolumn{6}{c}{\{256$\times$192, 384$\times$288\}}                     \\ \hline
\end{tabular}
}
\end{table*}

On the other hand, \textit{self-knowledge distillation} focuses on regularizing \textit{the knowledge itself}, that is, by forcing the deep learning to produce more meaningful and consistent predictions, \eg, in a class-wise manner~\cite{yun2020regularizing}. This technique also enables better knowledge transition in learning, such as in~\cite{kim2021self} that proposed a progressive self-knowledge distillation (PS-KD) to distill the model's knowledge as a soft target. MixSKD~\cite{yang2022mixskd} utilized two techniques, i.e., image mixture and self-knowledge distillation to mutually distill knowledge between random pair of images. A simple dropout sampling can be utilized to distill the posterior distribution of multiple models~\cite{lee2023self}. In the same light, our Self-KD regularizes the knowledge itself for better generalization. Yet, unlike these works, our technique chooses the teacher model with incremental performance via cross-validation. This can be done without computing the teacher model's gradients; thus, our technique requires only an additional feed-forward pass when computing back-propagation for the student model.

\section{Proposed Method}

We proposed a new framework for saliency prediction by incorporating two methods: \textit{(i) SalNAS}, a supernet for saliency prediction, and \textit{(ii) Self-KD}, a teacherless distillation. The proposed architecture of SalNAS consists of an encoder-decoder structure, as shown in Fig~\ref{fig:1}. It utilizes a search space derived from AlphaNet~\cite{wang2021alphanet} and is adjusted specifically for saliency prediction.  Then, we propose a self-knowledge distillation, termed Self-KD, to improve the generalization of SalNAS.  To train the SalNAS supernet, a combination loss that incorporates different saliency evaluation metrics is presented. SalNAS is first presented in Section~\ref{Sec:SalNAS}.  Self-KD is then proposed in Section~\ref{Sec:selfKD}. Then, the learning loss is discussed in Section~\ref{Sec:Loss}.

\subsection{SalNAS}
\label{Sec:SalNAS}
\subsubsection{Formulation}
\label{Sec:SalNAS:Formulation}
Let $\textit{Sal}_i$ denote the architectural configurations of subnets in a supernet to be used for saliency prediction. Let $W_o$ be the weight-sharing of the supernet. Our strategy is to minimize the validation loss function $\mathcal{L}_{val}$ by solving the optimization problem:

\begin{equation}
\label{eq:Opt}
  \min_{W_o}  \sum_{\textit{Sal}_i} \mathcal{L}_{val}(\mathcal{S}(W_o, \textit{Sal}_i))  
\end{equation}

The architecture for each $\textit{Sal}_i$ is a \acrshort{fcn} consisting of the dynamic encoder and decoder module. Specifically, we consider $\textit{Sal}_i := \textbf{Decoder}(\textbf{Encoder}(\mathcal{X}))$. $\mathcal{S}$ is a selection scheme that selects part of the model from the supernet to form a sub-network with architectural configuration $\textit{Sal}_i$. 

The dynamic convolutional layers can adapt to variable dimensions such as width, depth, kernel size, and input resolutions, unlike the vanilla convolutional layers with fixed feature and kernel sizes \cite{li2021dynamic, li2022ds, cai2019once}. Suppose we have an elastic input with shape $\mathcal{X} \in \mathbb{R}^{{C}'_{in} \times H \times W}$, where ${C}'_{in}$ denote the selected input channel; $H$ and $W$ denote the height and width of the elastic input, respectively. The dynamic convolutional layers are defined as follows:

\begin{equation}
\label{eq:1}
\psi_{conv}\left ( \mathcal{X} \right ) = \mathcal{W}\left [: {C}'_{out}, : {C}'_{in}  \right ] \odot \mathcal{X}
\end{equation}

\noindent \noindent where $\left [ ~:~ \right ]$ is a slice operation denoted in a Python-like style; $\mathcal{W} \in \mathbb{R}^{C_{out} \times C_{in} \times K \times K}$ denote the parameter of a convolutional layer; $C_{in}$, $C_{out}$, ${C}'_{out}$, and $K$ represent the input channel number, output channel number, selected output channel number, and kernel size.

Specifically, the encoder has dynamic layers with four variable dimensions, \ie, resolution, depth, expansion ratio, and kernel size. Each layer in the proposed dynamic encoder contains an efficient, dynamic MobileNet block (MB) based on MobileNetV3~\cite{howard2019searching}. Each dynamic MB block consists of dynamic point-wise convolution, dynamic depthwise separable convolution, squeeze-and-excite, and dynamic pointwise convolution. Then the decoder employs dynamic convolutional layers where a \acrshort{nas}-based approach is used to regenerate the output from the encoder to match the input resolution stage-wise, as shown in Fig.~\ref{fig:1}. 

The dynamic convolution in the proposed supernet enables efficient architecture searching. That is, our supernet can contain subnets with various configurations (\eg~sizes and weight-sharing) that can be trained as a single supernet and optimized for different hardware platforms \textit{without retraining and modifying the network's weights}. This reduces the computational time needed for architecture search and allows efficient inference across multiple devices.

\subsubsection{Architecture space}
\label{Sec:SalNAS:space}
The proposed method utilized AlphaNet~\cite{wang2021alphanet} trained on ImageNet for the encoder part of the \acrshort{fcn} network. AlphaNet is based on an efficient neural network design. It uses weight-sharing \acrshort{nas} to build a supernet (which contains many architectures as its subnets) and jointly trains a supernet and its subnets. AlphaNet utilizes an adaptive selection of alpha-divergence to avoid over or underestimating teacher models in a \acrshort{kd} framework, inplace \acrshort{kd}. We selected over $10^{19}$ subnets from AlphaNet supernet and modified it for the saliency prediction encoder. We utilized the search space dimension of AlphaNet. A summary of our proposed search space dimension is shown in Table~\ref{tab:1}. We have three blocks: Standard Convolution (Conv), MobileNetV3 convolution (MBConv), and Input resolution. The reason for selecting MBConv block is to maintain consistency with the search space of AlphaNet. Moreover, changing the encoder architecture would require re-training on ImageNet dataet. Each MBConv block has various widths, depths, kernels, and expansion ratios. For instance, the smallest subnet has the lowest computational complexity and parameters (0.51 billion FLOPS and 4.97 million parameters), while the largest subnet with the highest computational complexity and parameters (4.18 billion FLOPS and 19.38 million parameters). Furthermore, the decoder part of the proposed method adaptively adjusts the decoder module according to the equivalent encoder without human intervention. 

\subsection{Self-knowledge distillation}
\label{Sec:selfKD}

Traditional KD methods are challenging due to extensive testing for pairing teacher and student models. Teacherless KD minimizes this by using the student model from the previous epoch or mini-batch as the teacher for the current one, eliminating the need for a separate teacher model. Although it may not always outperform teacher-based methods, it significantly reduces model pairing and training time.

To enhance the generalization performance, we train the model in a \acrshort{kd} framework, consisting of a teacher model and a student model. Unlike the traditional \acrshort{kd} whose optimizer needs parameters from both models, our proposed Self-KD is a \textit{teacherless} \acrshort{kd}. That is, inspired by Stochastic Weight Averaging (SWA) \cite{Izmailov2018876}, Self-KD improves model generalization by averaging parameters at each epoch, stabilizing training, and improving performance on unseen datasets. Our teacher model is an averaged network whose weights are the average between the best-performing models (local minima) chosen via cross-validation. This results in better generalization---due to the use of validation data---as well as lower computation---since our approach does not need the gradient computation for the teacher model. For an input $x$ and the output ($p_{avg}$) from the teacher model $S_{avg}\left ( x \right )$ is used to augment the ground truth ($gt$) as follows:

\begin{equation}
\label{eq:3}
\bar{gt} = p_{avg} + (gt - p_{avg}) \times \alpha
\end{equation}

\noindent where $p_{avg}$ represents the augmented information from the averaged network's prediction; alpha ($\alpha$)  weights the belief towards the augmented information versus ground truth. Note that we set $\alpha = 1$ for the first epoch to omit the augment information; after the first epoch, $\alpha$ can be set to an appropriate value $\in (0,1)$ for the calculation in Eq.~\ref{eq:3}.

\begin{figure}[t!]
\begin{center}
\includegraphics[width=0.97\linewidth]{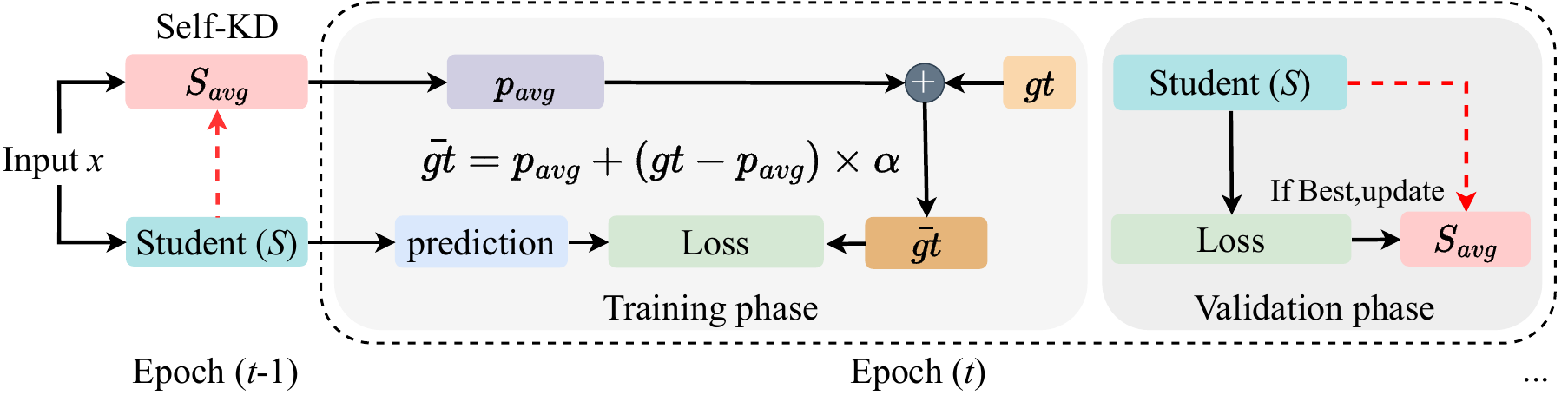}
\end{center}
\caption{The proposed Self-Knowledge Distillation method.}
\label{fig:selfkd}
\end{figure}

During the validation phase, the averaged model $S_{avg}$ is updated, if the validation loss is lower than the previous one. The update is done by averaging the model parameters with the previously chosen parameters of the best-performing student model $S$. Otherwise, the parameters of the averaged model $S_{avg}$ are retained. This process is illustrated in Fig.~\ref{fig:selfkd} and described in Algorithm~\ref{alg:1}. Although our approach requires slightly more computation time than the baseline training without distillation, it requires less training time than \acrshort{pkd} which involves the same teacher and student network. Additionally, Self-KD involves fewer networks compared to \acrshort{pkd}.

\begin{algorithm}[t!]
    \caption{Self-Knowledge Distillation}
    \label{alg:1}
    \begin{algorithmic}[1]
        \STATE {\bfseries Input:} Initialized Model $S$
        \STATE {\bfseries Output:} Trained Model $S$
        \STATE $S_{avg}$ is the averaged network, $p_{avg}$ is the prediction of averaged network, $p_s$ is the prediction of student network, $gt$ is the corresponding ground truth, $\mathcal{L}_{(\cdot)}$ is the associated loss functions, $D$ is the dataset, $E$ is number of epoch, $\alpha$ is 0.5.
        \FOR {each epoch in $E$}
            \FOR {each batch in $D_{train}$}
                \IF{epoch $>$ 1}
                    \STATE $p_{avg} \gets$ Forward $S_{avg}$ given input images;
                    \STATE $\bar{gt} \gets p_{avg} + (gt - p_{avg}) \times \alpha$;
                \ENDIF
                \STATE $p_{s} \gets$ Forward $S$ given input images;
                \IF{epoch $>$ 1}
                    \STATE Calculate $\mathcal{L}_{train}$ with $p_s, \bar{gt}$; (see Eq.~\ref{eq:9})
                \ELSE
                    \STATE Calculate $\mathcal{L}_{train}$ with $p_s, gt$; (see Eq.~\ref{eq:9})
                \ENDIF
                \STATE $S \gets$ Backward by $\mathcal{L}_{train}$; 
            \ENDFOR
            \STATE $\mathcal{L}_{val} \gets$ Evaluate $S$ on $D_{val}$;
            \IF{$\mathcal{L}_{val}$ is best value}
                \STATE $S_{avg}$ $\gets$ update parameters with $S$;
            \ENDIF
        \ENDFOR
        
        \STATE Return $S$
    \end{algorithmic}
\end{algorithm}

\subsection{Loss function}
\label{Sec:Loss}
We used three popular saliency prediction metrics to construct the learning loss functions, namely, \acrshort{kld}, \acrshort{cc}, and \acrshort{nss}. The \acrshort{kld} and \acrshort{cc} correspond to distribution-based metrics for saliency prediction; meanwhile, \acrshort{nss} corresponds to location-based metrics of saliency prediction~\cite{bylinskii2018different}. Our learning loss function is defined as follows:
\begin{equation}
\label{eq:9}
\mathcal{L}_{train} = KLD(P,G) + CC(P,G) + NSS(P,F)
\end{equation} 

\noindent where the loss formulations related to \acrshort{kld}, \acrshort{cc}, and \acrshort{nss} are denoted as $KLD(P,G)$, $CC(P,G)$, and $NSS(P,F)$, respectively. These losses compare the difference between the prediction $P$ and the related ground truth values $G$, as well as the fixation locations $F$.

Employing evaluation metrics in formulating the loss function is intuitive and is a common practice. Nevertheless, the range of these values is different by definition according to to~\cite{bylinskii2018different}, \eg,  $KLD(P,G)$ and $CC(P,G)$ can take the value between 0 and 1, but $NSS(P,F)$ can take any real, negative value. Therefore, we propose to adjust each evaluation metric to be employed as a loss function, described in the following.\\

$KLD(P,G)$ loss measures the divergence at each pixel location, it is defined as follows: 
\begin{equation}
\label{eq:4}
KLD(P,G) = \sum_{i}^{} G_{i} \log\left(\epsilon + \frac{G_{i}}{P_{i} + \epsilon}\right)
\end{equation} 

\noindent where  $G_{i}$ and $P_{i}$ denote the ground truth and the predicted saliency value  at $i^{th}$ pixel location; $\epsilon$ is a small constant to prevent numerical instability.

$CC(P,G)$ loss measures the correlation between the predicted and the ground truth saliency map $P$ and $G$, is defined as:

\begin{equation}
\label{eq:5}
CC(P,G) = 1 - \frac{{\text{covar}(P,G)}}{{\sigma(P) \times \sigma(G)}}
\end{equation}

\noindent where the covariance between $P$ and $G$ $\text{covar}(P,G)$ is normalized by the standard deviations of $P$ and $G$, denoted as $\sigma(P)$ and $\sigma(G)$,  respectively.

$NSS(P,F)$ loss is computed as the average normalized saliency at a fixated location. Our \acrshort{nss} loss is defined as

\begin{equation}
\label{eq:6}
NSS(P,F) = 1 - \left ( \frac{1}{N} \sum_{i}^{} \sigma(\bar P_{i} \times F_{i}) \right )
\end{equation}

\noindent where  $\bar P$ represents the normalized saliency map $P$ with a mean of zero and a standard deviation of one; $F$ denotes the binary map of fixation locations. $N$ denotes the total number of human eye fixations. We scale NSS to the range of 0 to 1 by using a logistic function $\sigma$. The logistic function $\sigma$ is defined as:

\begin{equation}
\label{eq:7}
\sigma(x) = \frac{1}{1 + e^{-x}}
\end{equation}

\noindent This adjustment makes the metric more usable and interpretable and ensures the loss function is differentiable.

\section{Experimental Results}

The baseline models used in our experiments, which can be categorized as \textit{compact}, \textit{deep}, and \textit{hybrid} for a fair comparison, are introduced in Section~\ref{Sec:baseline}. The detailed implementation of our work and the settings during training are described in Section~\ref{Sec:Implementation} and~\ref{Sec:LR}. Our evaluation is performed across several practical scenarios with the protocols and datasets provided in Section~\ref{Sec:Protocal}. The performance analysis against the state-of-the-art saliency prediction is in Section~\ref{Sec:Perform}, while the comparison against state-of-the-art backbones is analyzed in the transfer learning setting Section~\ref{Sec:Transfer}. The qualitative analysis is provided in Section~\ref{Sec:Qualitative}, and the real-time performance is provided in Section~\ref{Sec:Realtime}.

\subsection{Baseline}
\label{Sec:baseline}
 
In the following experiments, we divided the baseline models into three categories, for fair comparison: \textit{compact}, \textit{deep}, and \textit{hybrid} models.   \textit{Compact model} category refers to the small and efficient models; \textit{deep model} category refers to the large-size models, while \textit{hybrid model} category refers to the group of models with the combination of architectures, e.g., CNN and Vision transformer.  These baseline models are used as references in performance analysis and the real-time processing saliency prediction in Section~\ref{Sec:Perform} and~\ref{Sec:Realtime}, respectively.

In performance analysis Section~\ref{Sec:Perform}, we employed the following baseline models to compare against state-of-the-art saliency prediction. This selection is driven by the availability of SALICON test set results and source code. \textit{The compact model} is FastSal \cite{hu2021fastsal}; the \textit{deep model}  includes SimpleNet \cite{reddy2020tidying}, EML-NET \cite{jia2020eml}, MSI-Net \cite{kroner2020contextual}, SalFBNet \cite{ding2022salfbnet}, DINet \cite{yang2019dilated}, SalED \cite{wang2021saled}, DeepGaze II-E \cite{linardos2021deepgaze}, ACSalNet \cite{qing2022attentive}, FastFixation \cite{liang2023fastsal}, SalDA \cite{tang2023salda}, and TempSal \cite{aydemir2023tempsal}; the \textit{hybrid model} includes TranSalNet \cite{lou2022transalnet}. \\
 
For the analysis of the performance gained by Self-KD training, we employed the following baseline models, according to the above categorization: the~\textit{compact model} include EEEA-Net-C2 \cite{termritthikun2021eeea}, OFA595 \cite{cai2019once}, MobileNetV3 \cite{howard2019searching}, and the~\textit{deep model} includes EfficientNet-B4 \cite{tan2019efficientnet} and TResNet-M \cite{ridnik2021tresnet}. We employ a similar selection for the architecture transfer section.  \\

For real-time processing performance Section~\ref{Sec:Realtime}, state-of-the-art saliency prediction models are considered. The baseline models, which can be put into the above categorization, are as follows: the~\textit{compact model}   includes EEEA-Net-C2, OFA595, and EfficientNet-B0, the \textit{deep model}   includes SimpleNet, SalFBNet, EfficientNet-B4, EfficientNet-B7, while the \textit{hybrid model}   includes TranSalNet.

\subsection{Implementation details}
\label{Sec:Implementation}
Our proposed method is implemented using PyTorch. We utilized AlphaNet search space for the encoder module for training supernet, while the decoder automatically adjusted dimensions according to the encoder. Furthermore, we used mixed precision during supernet training with a maximum input resolution size of $384 \times 288$. For hyperparameter configuration, our approach aligns with the setting of AlphaNet. SGD optimizer is utilized with a momentum of 0.9, a learning rate (LR) of 0.1, the LR scheduler is cosine annealing, and batch size is reduced to 64 in accordance with our hardware memory specifications. Supernet is first trained on the SALICON dataset \cite{jiang2015salicon} and then fine-tuned on other datasets such as CAT2000 \cite{kummererSaliencyBenchmarkingMade2018}, MIT1003 \cite{kummererSaliencyBenchmarkingMade2018}, OSIE \cite{xu2014predicting}, DUT-OMRON \cite{yang2013saliency}, PASCAL-S \cite{li2014secrets}, and FIWI \cite{shen2014webpage}.

\subsection{LR scheduler and Combination loss} 
\label{Sec:LR}

\textbf{LR scheduler.}
We utilized a cosine annealing LR scheduler, the learning rate starts at the initial value and decreases to a minimum value following a cosine annealing schedule. $T_0$ is used to recalibrate the learning rate to its highest value, which allows the model to avoid getting stuck in local minima, thus allowing the model to continue learning and improving performance. Table~\ref{tab:3} shows the improved performance due to a varying $T_0$. This indicates that the model performs better when it is trained with a varied epoch and $T_0$ values. The configuration remained consistent with $T_0$ set to 10 and epochs set to 20 in all experiments.

\textbf{Combination loss analysis.} The performance of a model can be enhanced by combining two or more losses. We found that effective performance may be achieved by adjusting these losses with similar ranges. Table~\ref{tab:4} illustrates the effectiveness of different combinations, and the combined loss $CC+KLD+NSS$ offers a significant improvement in the \acrshort{nss} and an improvement in \acrshort{auc} metrics. Thus, we utilized this combination loss in all our experiments.

\begin{table}[t!]
\centering
\caption{Overview of Saliency Prediction Datasets: Training and Validation Data, Image Resolutions.}
\resizebox{\columnwidth}{!}{%
\label{tab:datasets}
\begin{tabular}{lccc}
\hline
\textbf{Dataset} & \textbf{\#Training} & \textbf{\#Validation } & \textbf{Image resolution} \\ \hline
SALICON & 10000 & 5000 & 640 $\times$ 480 \\
CAT2000 & 1600 & 400 & 1920 $\times$ 1080 \\
MIT1003 & 800 & 200 & 1280 $\times$ 1024 \\
PASCAL-S & 650 & 200 & 500 $\times$ 375 \\
DUT-OMRON & 4168 & 100 & 400 $\times$ 400 \\
OSIE & 500 & 200 & 800 $\times$ 600 \\
FIWI & 99 & 50 & 1360 $\times$ 768 \\  \hline
\end{tabular}
}
\end{table}

\begin{table}[t!]
\centering
\caption{Performance Evaluation of SalNAS with Different Training Epochs and $T_0$ Values. ↑ Represents Higher Values (Better), and ↓ Represents Lower Values (Better).}
\label{tab:3}
\resizebox{\columnwidth}{!}{%
\begin{tabular}{ccccccc}
\hline
\textbf{Epoch} & $\mathbf{T_{0}}$ & \textbf{CC} $\uparrow$ & \textbf{KLD} $\downarrow$ & \textbf{NSS} $\uparrow$ & \textbf{SIM} $\uparrow$ & \textbf{AUC} $\uparrow$ \\ \hline
10 & 10 & 0.9080 & 0.1921 & 1.9757 & \textbf{0.8012} & 0.8595 \\
20 & 20 & 0.9067 & 0.1923 & 1.9747 & 0.7996 & 0.8595 \\
30 & 30 & 0.9073 & 0.1945 & 1.9674 & 0.8002 & 0.8594 \\
20 & 10 & \textbf{0.9082} & 0.1921 & \textbf{1.9764} & \textbf{0.8012} & 0.8594 \\
30 & 15 & 0.9078 & \textbf{0.1913} & 1.9748 & 0.8008 & \textbf{0.8598} \\
30 & 10 & \textbf{0.9082} & 0.1919 & 1.9762 & \textbf{0.8012} & 0.8596 \\ \hline
\end{tabular}
}
\end{table}

\begin{table}[t!]
\centering
\caption{Performance Analysis of SalNAS with Different Loss Combinations.}
\label{tab:4}
\resizebox{\columnwidth}{!}{%
\begin{tabular}{lccccc}
\hline
\textbf{Combination} & \textbf{CC} $\uparrow$ & \textbf{KLD} $\downarrow$ & \textbf{NSS} $\uparrow$ & \textbf{SIM} $\uparrow$ & \textbf{AUC} $\uparrow$ \\ \hline
\acrshort{pkd} Loss & 0.9082 & \textbf{0.1910} & 1.9554 & 0.8003 & 0.8591 \\ \hline
CC + KLD & 0.9080 & \textbf{0.1910} & 1.9548 & 0.8002 & 0.8590 \\
CC + KLD + SIM & \textbf{0.9083} & 0.1961 & 1.9560 & \textbf{0.8012} & 0.8590 \\
CC + KLD + NSS & 0.9079 & 0.1957 & \textbf{1.9733} & 0.8009 & \textbf{0.8594} \\
CC + KLD + AUC & 0.9082 & 0.1910 & 1.9556 & 0.8003 & 0.8592 \\ \hline
\end{tabular}%
}
\end{table}

\subsection{Evaluation protocols} 
\label{Sec:Protocal}
\noindent \textbf{Datasets.} We considered seven publicly available saliency prediction datasets for evaluating saliency prediction models: SALICON, MIT1003, CAT2000, PASCAL-S, DUT-OMRON, OSIE, and FIWI. Table~\ref{tab:datasets} shows detailed information about these datasets.

\noindent \textbf{Evaluation metrics.} 
The evaluation of saliency prediction models is based on the consistency between predicted saliency and the ground truth. Following~\cite{bylinskii2018different}, we considered five widely accepted evaluation metrics: \acrshort{nss}, \acrshort{cc}, \acrshort{kld}, \acrshort{sim}, and \acrshort{auc}. \acrshort{kld}, \acrshort{sim}, and \acrshort{cc} are distribution-based metrics, while \acrshort{auc} and \acrshort{nss} are location-based metrics \cite{bylinskii2018different}.

The \acrshort{cc} metric is a statistical method to measure the correlation between two dependent variables, its range is between -1 to 1, where value near 1 represents high correlation between two saliency maps. The \acrshort{kld} metric is utilized to assess dissimilarity between two probability distributions, where a lower value represents a better alignment between ground truth and predicted saliency map. \acrshort{nss} metric calculates an average normalized saliency value obtained at the fixation locations of the normalized saliency map. \acrshort{sim} metric is widely utilized for saliency prediction by measuring the similarity between two saliency maps (viewed as histograms). \acrshort{auc} metric treats the saliency map as a binary classifier of fixations (to separate the positive and negative samples) at different threshold levels.


\begin{table}[!t]
\centering
\caption{Quantitative Performance Comparison on SALICON Benchmark (Test set): The Top Three Results are Highlighted in \textbf{{\color[HTML]{FE0000} Red}}, \textbf{{\color[HTML]{3166FF} Blue}}, and \textbf{{\color[HTML]{32CB00} Green}}, respectively.}
\resizebox{\columnwidth}{!}{%
\label{tab:5}
\begin{tabular}{lccccccc}
\hline
 & \multicolumn{7}{c}{\textbf{SALICON benchmark}} \\ \cline{2-8} 
\textbf{Model} & \textbf{sAUC} $\uparrow$ & \textbf{IG} $\uparrow$ & \textbf{NSS} $\uparrow$ & \textbf{CC} $\uparrow$ & \textbf{AUC} $\uparrow$ & \textbf{SIM} $\uparrow$ & \textbf{KLD} $\downarrow$ \\ \hline
SimpleNet & 0.743 & 0.880 & 1.960 & 0.907 & {\color[HTML]{3166FF} \textbf{0.869}} & 0.793 & 0.201 \\
EML-NET & 0.746 & 0.736 & {\color[HTML]{FE0000} \textbf{2.050}} & 0.886 & 0.866 & 0.780 & 0.520 \\
MSI-Net & 0.736 & 0.793 & 1.931 & 0.889 & 0.865 & 0.784 & 0.307 \\
SalFBNet & 0.740 & 0.839 & 1.952 & 0.892 & {\color[HTML]{32CB00} \textbf{0.868}} & 0.772 & 0.236 \\
FastSal & 0.732 & 0.770 & 1.845 & 0.874 & 0.863 & 0.768 & 0.288 \\
DINet & 0.739 & 0.195 & 1.959 & 0.902 & 0.862 & 0.795 & 0.864 \\
SalED & 0.745 & {\color[HTML]{3166FF} \textbf{0.909}} & 1.984 & {\color[HTML]{3166FF} \textbf{0.910}} & {\color[HTML]{3166FF} \textbf{0.869}} & {\color[HTML]{3166FF} \textbf{0.801}} & {\color[HTML]{FE0000} \textbf{0.190}} \\
DeepGaze II-E & {\color[HTML]{FE0000} \textbf{0.767}} & 0.766 & 1.996 & 0.872 & {\color[HTML]{3166FF} \textbf{0.869}} & 0.733 & 0.285 \\
ACSalNet & 0.744 & 0.890 & 1.981 & 0.905 & {\color[HTML]{32CB00} \textbf{0.868}} & 0.798 & 0.232 \\
SalDA & 0.714 & 0.577 & 1.727 & 0.821 & 0.851 & 0.693 & 0.369 \\
FastFixation & 0.736 & 0.687 & 1.901 & 0.879 & 0.863 & 0.766 & 0.407 \\
TempSal & 0.745 & 0.896 & 1.967 & {\color[HTML]{FE0000} \textbf{0.911}} & {\color[HTML]{3166FF} \textbf{0.869}} & {\color[HTML]{32CB00} \textbf{0.800}} & {\color[HTML]{3166FF} \textbf{0.195}} \\
TranSalNet & {\color[HTML]{32CB00} \textbf{0.747}} & - & {\color[HTML]{32CB00} \textbf{2.014}} & 0.907 & {\color[HTML]{32CB00} \textbf{0.868}} & {\color[HTML]{FE0000} \textbf{0.803}} & 0.373 \\ \hline
SalNAS-XL (ours) & {\color[HTML]{32CB00} \textbf{0.747}} & {\color[HTML]{32CB00} \textbf{0.902}} & 2.004 & 0.907 & {\color[HTML]{3166FF} \textbf{0.869}} & 0.794 & {\color[HTML]{32CB00} \textbf{0.200}} \\
SalNAS-XL + Self-KD & {\color[HTML]{3166FF} \textbf{0.749}} & {\color[HTML]{FE0000} \textbf{0.913}} & {\color[HTML]{3166FF} \textbf{2.019}} & {\color[HTML]{32CB00} \textbf{0.909}} & {\color[HTML]{FE0000} \textbf{0.870}} & 0.796 & {\color[HTML]{3166FF} \textbf{0.195}} \\ \hline
\end{tabular}%
}
\end{table}

\begin{figure*}
    \centering
    
    \begin{subfigure}{0.5\textwidth}
        \centering
        \includegraphics[width=1\linewidth]{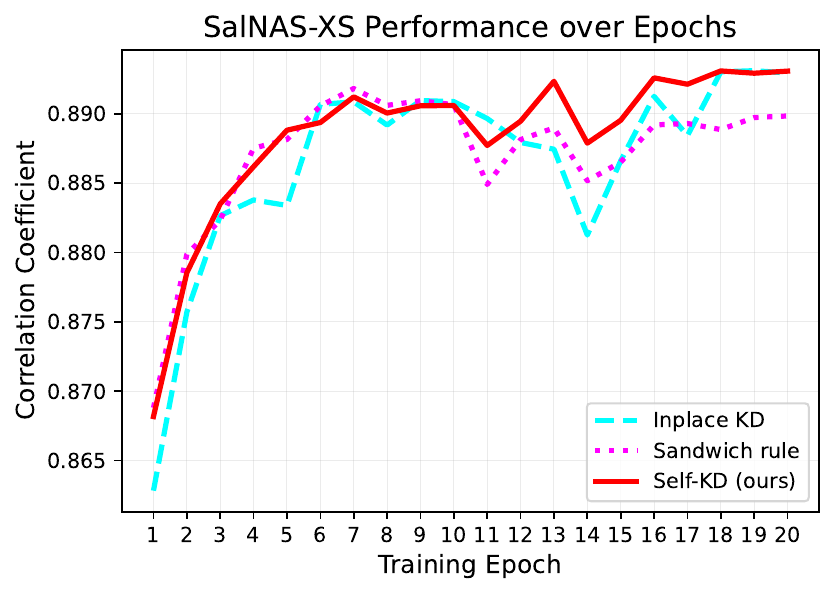}
        \caption{The smallest subnet (SalNAS-XS)}
        \label{fig:image1}
    \end{subfigure}%
    \begin{subfigure}{0.5\textwidth}
        \centering
        \includegraphics[width=1\linewidth]{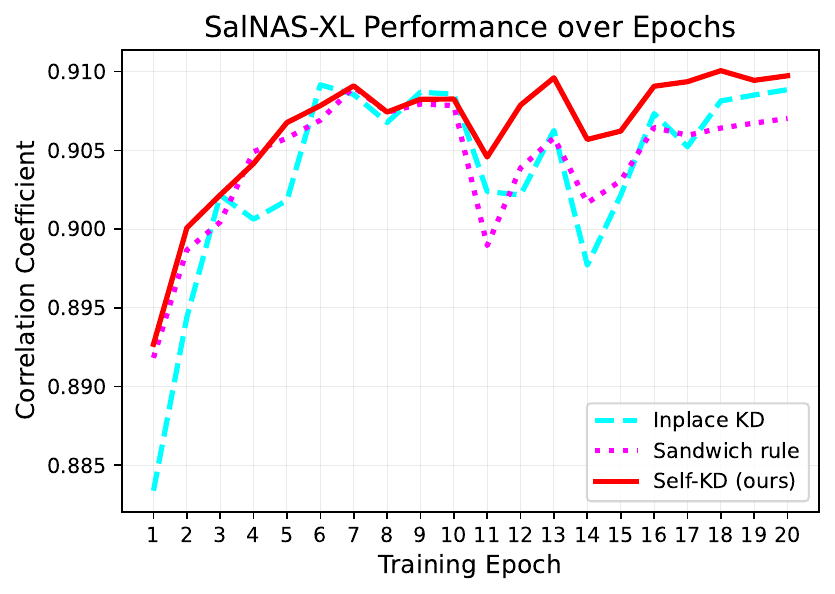}
        \caption{The largest subnet (SalNAS-XL)}
        \label{fig:image2}
    \end{subfigure}
    
    \caption{(\ref{fig:image1}) shows the correlation coefficient for the smallest subnet (SalNAS-XS), showing superior performance with the Self-KD method. (\ref{fig:image2}) shows the correlation coefficient for the largest subnet (SalNAS-XL) shows the Self-KD method's superiority over inplace distillation and the sandwich rule methods.}
    \label{fig:epoch_cc}
\end{figure*}

\subsection{Performance analysis} 

\label{Sec:Perform}

\textbf{Comparison against the state-of-the-art saliency prediction.}  We trained our SalNAS on the SALICON training set and submitted the results to the SALICON benchmark\footnote{\url{http://salicon.net/challenge-2017/}.}, which is a saliency prediction challenge. The goal of the challenge is to evaluate the performance of visual saliency prediction of natural scene images. The SALICON test dataset comprises 5000 images with no ground truths released. Our SalNAS-XL is a subnet with the maximum computational complexity, parameters, and performance from our proposed supernet SalNAS. We compared the result of SalNAS-XL with thirteen recent state-of-the-art saliency prediction models, including SimpleNet,  EML-NET, MSI-Net, SalFBNet, FastSal, DINet, SalED, DeepGaze II-E, ACSalNet, FastFixation, SalDA, TempSal, and TranSalNet. Table~\ref{tab:5} provides the result on the benchmark, which shows that SalNAS-XL has competitive results\footnote{\url{https://codalab.lisn.upsaclay.fr/competitions/8379}.} compared to other state-of-the-art. SalNAS-XL with Self-KD has achieved the best performance, \ie, it gives the highest IG and \acrshort{auc} and provides the second best results with minor differences to the highest ones, \ie,  2.3\%, 1.5\%, and 2.5\% on sAUC, \acrshort{nss}, and \acrshort{kld}, respectively.  Our SalNAS-XL without Self-KD offers competitive performance, \ie, it gives the third-best sAUC, IG, and \acrshort{kld}, while giving the second-best \acrshort{auc} among other state-of-the-art.

From Table.~\ref{tab:5}, it can be observed that few saliency models, such as DeepGaze II-E, EML-NET, SalED, TempSal, and TransSalNet, outperformed SalNAS in various metrics: sAUC, \acrshort{nss}, \acrshort{cc}, \acrshort{kld}, and \acrshort{sim}. One reason for the enhanced performance of these state-of-the-art models is the utilization of ResNet-50 and DenseNet backbones, which have significantly high model parameters ranging from 23 to 104 million, compared with our SalNAS with 20.98 million parameters.

\textbf{Analysis of performance gain by Self-KD training.} We compared Self-KD against the non-distillation (baseline), \acrfull{pkd} \cite{umer2022device}, and PS-KD training \cite{kim2021self}. We show the performance across different backbones, \ie, our SalNAS-XL and five other backbones from state-of-the-art saliency prediction---EEEA-Net-C2, OFA595, MobileNetV3, EfficientNet-B4, and TResNet-M. These backbones can be separated into \textit{(i) the small-size backbones} with the number of parameters $< 10M$ and \textit{(ii) the medium-to-large-size backbones} with the number of parameters $> 20M$.

From Table~\ref{tab:6}, Self-KD outperformed the baseline, \acrshort{pkd}, and PS-KD training for both \textit{the small-size} and \textit{the medium-to-large-size backbones}, across all metrics, except for \acrshort{nss}. In addition to superior performance, Self-KD requires less training time compared to \acrshort{pkd}. Furthermore, compared to the baseline training, Self-KD only requires 10\% higher training time and$~$10-20\% higher training time for \textit{the small-size} and \textit{the medium-to-large-size backbones}, respectively.

We explored other efficient training strategies, for instance, sandwich rule and inplace distillation to validate the performance of the proposed SalNAS method. In the sandwich rule, each mini-batch of data samples smallest subnet, biggest subnet, and two randomly sampled subnet models. The gradient from all sampled subnet models is aggregated before updating the weights of the single-stage model. The inplace distillation involves soft labels generated by the biggest possible subnet model to supervise all other subnet models. Fig.~\ref{fig:epoch_cc} illustrates the performance of the proposed Self-KD method compared to inplace distillation and the sandwich rule methods, with the \acrshort{cc} metric considered for evaluation. From the results, it can be observed that the Self-KD method outperforms other efficient training strategies including Inplace distillation and sandwich rule.

Furthermore, comprehensive hyperparameter tuning experiments are conducted to explore the impact of varying $\alpha$ values on the model performance across diverse saliency datasets. The results show that the $\alpha$ value between 0.4 and 0.6 gives good performance across different datasets.

\subsection{Architecture transfer}
\label{Sec:Transfer}
This section evaluates the performance of the pre-trained backbones after fine-tuning with six different datasets, namely MIT1003, CAT2000, PASCAL-S, DUT-OMRON, OSIE, and FIWI datasets. We compared results with state-of-the-art backbones, as shown in Table~\ref{tab:7} and Table~\ref{tab:8}. All the backbones were pre-trained with the SALICON training dataset. For our work, we used a pre-trained SALICON subnet SalNAS-XL from our proposed supernet and then fine-tuned it. From the table,  our SalNAS-XL wins 25 times out of 30 trials (six datasets evaluated over five metrics), confirming the effectiveness of our strategies. \\

\begin{table*}[t!]
\centering
\caption{Comparison between different \acrshort{kd} methods across architectures on SALICON validation dataset.}
\label{tab:6}
\resizebox{\textwidth}{!}{%
\begin{tabular}{lclccccccrc}
\hline
\multicolumn{1}{c}{\textbf{Method}} & \textbf{Teacher} & \textbf{Student} & \textbf{CC} $\uparrow$  & \textbf{KLD} $\downarrow$ & \textbf{NSS} $\uparrow$  & \textbf{SIM} $\uparrow$  & \textbf{AUC} $\uparrow$  & \textbf{Time (min)} & \textbf{ Change} & \textbf{Params (M)} \\ \hline
Baseline & - & EEEA-Net-C2& 0.8966 & 0.2072 & 1.9414 & 0.7891 & 0.8538 & 23 & 1$\times$ & 4.57 \\
PKD & EEEA-Net-C2& EEEA-Net-C2& 0.8973 & 0.2062 & \textbf{1.9622} & 0.7910 & 0.8543 & 31 & 1.3$\times$ & 9.14 \\
PS-KD & - & EEEA-Net-C2& 0.8971 & 0.2044 & 1.9387 & 0.7895 & 0.8547 & 25 & 1.1$\times$ & 4.57 \\
Self-KD (ours) & - & EEEA-Net-C2& \textbf{0.9028} & \textbf{0.1952} & 1.9567 & \textbf{0.7937} & \textbf{0.8556} & 25 & 1.1$\times$ & 4.57 \\ \hline
Baseline & - & OFA595& 0.8964 & 0.2081 & 1.9340 & 0.7904 & 0.8550 & 24 & 1$\times$ & 7.41 \\
PKD & OFA595& OFA595& 0.8996 & 0.2018 & 1.9624 & 0.7934 & 0.8552 & 35 & 1.5$\times$ & 14.82 \\
PS-KD & - & OFA595& 0.8986 & 0.2082 & 1.9559 & 0.7927 & 0.8552 & 27 & 1.1$\times$ & 7.41 \\
Self-KD (ours) & - & OFA595& \textbf{0.9018} & \textbf{0.1968} & \textbf{1.9684} & \textbf{0.7952} & \textbf{0.8554} & 27 & 1.1$\times$ & 7.41 \\ \hline
Baseline & - & MobileNetV3& 0.8961 & 0.2124 & 1.9263 & 0.7899 & 0.8558 & 23 & 1$\times$ & 4.05 \\
PKD & MobileNetV3 & MobileNetV3& 0.8969 & 0.2048 & 1.9431 & 0.7899 & 0.8554 & 27 & 1.2$\times$ & 8.10 \\
PS-KD & - & MobileNetV3& 0.8971 & 0.2072 & 1.9278 & 0.7902 & \textbf{0.8562} & 23 & 1$\times$ & 4.05 \\
Self-KD (ours) & - & MobileNetV3& \textbf{0.9005} & \textbf{0.1987} & \textbf{1.9444} & \textbf{0.7924} & \textbf{0.8562} & 23 & 1$\times$ & 4.05 \\ \hline
Baseline & - & EfficientNet-B4& 0.9049 & 0.1987 & 1.9536 & 0.7977 & 0.8566 & 31 & 1$\times$ & 20.41 \\
PKD  & EfficientNet-B4& EfficientNet-B4& 0.9090 & 0.1869 & 1.9585 & 0.8021 & 0.8576 & 50 & 1.6$\times$ & 40.82 \\
PS-KD & - & EfficientNet-B4& 0.9068 & 0.1983 & 1.9645 & 0.8002 & 0.8580 & 37 & 1.2$\times$ & 20.41 \\
Self-KD (ours) & - & EfficientNet-B4& \textbf{0.9105} & \textbf{0.1839} & \textbf{1.9717} & \textbf{0.8029} & \textbf{0.8581} & 37 & 1.2$\times$ & 20.41 \\ \hline
Baseline & - & TResNet-M& 0.9036 & 0.1998 & 1.9476 & 0.7967 & 0.8560 & 25 & 1$\times$ & 52.96 \\
PKD  & TResNet-M& TResNet-M& 0.9072 & 0.1910 & \textbf{1.9746} & 0.8010 & 0.8562 & 36 & 1.4$\times$ & 105.92 \\
PS-KD & - & TResNet-M& 0.9057 & 0.1947 & 1.9575 & 0.7993 & 0.8553 & 30 & 1.2$\times$ & 52.96 \\
Self-KD (ours) & - & TResNet-M& \textbf{0.9100} & \textbf{0.1871} & 1.9649 & \textbf{0.8032} & \textbf{0.8565} & 30 & 1.2$\times$ & 52.96 \\ \hline
Baseline & - & SalNAS-XL (ours) & 0.9087 & 0.1908 & 1.9686 & 0.8015 & 0.8587 & 81 & 1$\times$ & 20.98 \\
PKD & SalNAS-XL (ours) & SalNAS-XL (ours) & 0.9090 & 0.1897 & \textbf{1.9889} & 0.8024 & 0.8594 & 100 & 1.2$\times$ & 41.96 \\
PS-KD & - & SalNAS-XL (ours) & 0.9088 & 0.1881 & 1.9714 & 0.8015 & 0.8592 & 90 & 1.1$\times$ & 20.98 \\
Self-KD (ours) & - & SalNAS-XL (ours) & \textbf{0.9117} & \textbf{0.1817} & 1.9789 & \textbf{0.8030} & \textbf{0.8603} & 90 & 1.1$\times$ & 20.98 \\ \hline
\end{tabular}
}
\end{table*}


\begin{table*}[t!]
\centering
\caption{Architecture transfer result on MIT1003, CAT2000, and PASCAL-S datasets.}
\label{tab:7}
\resizebox{\textwidth}{!}{%
\begin{tabular}{lccccc|ccccc|ccccc}
\hline
\multicolumn{1}{c}{\multirow{2}{*}{\textbf{Model}}} & \multicolumn{5}{c|}{\textbf{MIT1003}} & \multicolumn{5}{c|}{\textbf{CAT2000}} & \multicolumn{5}{c}{\textbf{PASCAL-S}} \\ \cline{2-16} 
\multicolumn{1}{c}{} & \multicolumn{1}{c}{\textbf{\begin{tabular}[c]{@{}c@{}}CC $\uparrow$\end{tabular}}} & \multicolumn{1}{c}{\textbf{\begin{tabular}[c]{@{}c@{}}KLD $\downarrow$\end{tabular}}} & \multicolumn{1}{c}{\textbf{\begin{tabular}[c]{@{}c@{}}NSS $\uparrow$\end{tabular}}} & \multicolumn{1}{c}{\textbf{\begin{tabular}[c]{@{}c@{}}SIM $\uparrow$\end{tabular}}} & \multicolumn{1}{c|}{\textbf{\begin{tabular}[c]{@{}c@{}}AUC $\uparrow$\end{tabular}}} & \multicolumn{1}{c}{\textbf{\begin{tabular}[c]{@{}c@{}}CC $\uparrow$\end{tabular}}} & \multicolumn{1}{c}{\textbf{\begin{tabular}[c]{@{}c@{}}KLD $\downarrow$\end{tabular}}} & \multicolumn{1}{c}{\textbf{\begin{tabular}[c]{@{}c@{}}NSS $\uparrow$\end{tabular}}} & \multicolumn{1}{c}{\textbf{\begin{tabular}[c]{@{}c@{}}SIM $\uparrow$\end{tabular}}} & \multicolumn{1}{c|}{\textbf{\begin{tabular}[c]{@{}c@{}}AUC $\uparrow$\end{tabular}}} & \textbf{\begin{tabular}[c]{@{}c@{}}CC $\uparrow$\end{tabular}} & \textbf{\begin{tabular}[c]{@{}c@{}}KLD $\downarrow$\end{tabular}} & \textbf{\begin{tabular}[c]{@{}c@{}}NSS $\uparrow$\end{tabular}} & \textbf{\begin{tabular}[c]{@{}c@{}}SIM $\uparrow$\end{tabular}} & \textbf{\begin{tabular}[c]{@{}c@{}}AUC $\uparrow$\end{tabular}} \\ \hline
EEEA-Net-C2 & 0.763 & 0.591 & 2.722 & 0.615 & 0.883 & 0.867 & 0.307 & 2.419 & 0.741 & \textbf{0.901} & 0.831 & 0.453 & 2.815 & 0.670 & 0.903 \\
OFA595 & 0.764 & 0.578 & 2.727 & 0.620 & 0.883 & 0.857 & 0.315 & 2.373 & 0.734 & 0.898 & 0.829 & 0.461 & 2.780 & 0.670 & 0.903 \\
MobileNetV3 & 0.749 & 0.606 & 2.677 & 0.611 & \textbf{0.891} & 0.861 & 0.322 & 2.395 & 0.734 & 0.893 & 0.817 & 0.493 & 2.783 & 0.659 & 0.901 \\
EfficientNet-B4 & 0.777 & 0.565 & 2.777 & 0.627 & 0.891 & 0.873 & 0.299 & 2.418 & 0.746 & 0.898 & 0.842 & 0.449 & 2.844 & \textbf{0.685} & \textbf{0.906} \\
TResNet-M  & 0.763 & 0.596 & 2.735 & 0.613 & 0.887 & 0.865 & 0.312 & 2.422 & 0.738 & 0.894 & 0.833 & 0.459 & 2.835 & 0.676 & 0.900 \\
SalNAS-XL (ours) & \textbf{0.794} & \textbf{0.528} & \textbf{2.892} & \textbf{0.641} & 0.879 & \textbf{0.877} & \textbf{0.288} & \textbf{2.462} & \textbf{0.752} & 0.896 & \textbf{0.846} & \textbf{0.425} & \textbf{2.879} & \textbf{0.685} & 0.905 \\ \hline
\end{tabular}%
}
\end{table*}


\begin{table*}[t!]
\centering
\caption{Architecture transfer result on DUT-OMRON, OSIE, and FIWI datasets.}
\label{tab:8}
\resizebox{\textwidth}{!}{%
\begin{tabular}{lccccc|ccccc|ccccc}
\hline
\multicolumn{1}{c}{\multirow{2}{*}{\textbf{Model}}} & \multicolumn{5}{c|}{\textbf{DUT-OMRON}} & \multicolumn{5}{c|}{\textbf{OSIE}} & \multicolumn{5}{c}{\textbf{FIWI}} \\ \cline{2-16} 
\multicolumn{1}{c}{} & \multicolumn{1}{c}{\textbf{\begin{tabular}[c]{@{}c@{}}CC $\uparrow$\end{tabular}}} & \multicolumn{1}{c}{\textbf{\begin{tabular}[c]{@{}c@{}}KLD $\downarrow$\end{tabular}}} & \multicolumn{1}{c}{\textbf{\begin{tabular}[c]{@{}c@{}}NSS $\uparrow$\end{tabular}}} & \multicolumn{1}{c}{\textbf{\begin{tabular}[c]{@{}c@{}}SIM $\uparrow$\end{tabular}}} & \multicolumn{1}{c|}{\textbf{\begin{tabular}[c]{@{}c@{}}AUC $\uparrow$\end{tabular}}} & \multicolumn{1}{c}{\textbf{\begin{tabular}[c]{@{}c@{}}CC $\uparrow$\end{tabular}}} & \multicolumn{1}{c}{\textbf{\begin{tabular}[c]{@{}c@{}}KLD $\downarrow$\end{tabular}}} & \multicolumn{1}{c}{\textbf{\begin{tabular}[c]{@{}c@{}}NSS $\uparrow$\end{tabular}}} & \multicolumn{1}{c}{\textbf{\begin{tabular}[c]{@{}c@{}}SIM $\uparrow$\end{tabular}}} & \multicolumn{1}{c|}{\textbf{\begin{tabular}[c]{@{}c@{}}AUC $\uparrow$\end{tabular}}} & \textbf{\begin{tabular}[c]{@{}c@{}}CC $\uparrow$\end{tabular}} & \textbf{\begin{tabular}[c]{@{}c@{}}KLD $\downarrow$\end{tabular}} & \textbf{\begin{tabular}[c]{@{}c@{}}NSS $\uparrow$\end{tabular}} & \textbf{\begin{tabular}[c]{@{}c@{}}SIM $\uparrow$\end{tabular}} & \textbf{\begin{tabular}[c]{@{}c@{}}AUC $\uparrow$\end{tabular}} \\ \hline
EEEA-Net-C2 & 0.785 & 0.496 & 3.301 & 0.653 & 0.917 & 0.816 & 0.501 & 3.470 & 0.644 & 0.948 & 0.765 & 0.385 & 1.718 & 0.676 & 0.844 \\
OFA595 & 0.781 & 0.503 & 3.302 & 0.652 & 0.917 & 0.811 & 0.518 & 3.442 & 0.636 & 0.948 & 0.745 & 0.412 & 1.679 & 0.664 & \textbf{0.852} \\
MobileNetV3 & 0.783 & 0.504 & 3.294 & 0.651 & 0.920 & 0.805 & 0.522 & 3.416 & 0.636 & 0.947 & 0.737 & 0.425 & 1.648 & 0.652 & 0.825 \\
EfficientNet-B4  & 0.793 & 0.483 & 3.334 & 0.657 & 0.919 & 0.838 & 0.450 & 3.557 & 0.671 & \textbf{0.949} & 0.781 & 0.366 & 1.770 & 0.688 & 0.846 \\
TResNet-M  & 0.781 & 0.506 & 3.282 & 0.653 & 0.915 & 0.829 & 0.479 & 3.493 & 0.648 & 0.948 & 0.611 & 0.602 & 1.283 & 0.592 & 0.810 \\
SalNAS-XL (ours) & \textbf{0.796} & \textbf{0.474} & \textbf{3.348} & \textbf{0.663} & \textbf{0.920} & \textbf{0.852} & \textbf{0.429} & \textbf{3.672} & \textbf{0.681} & 0.946 & \textbf{0.789} & \textbf{0.362} & \textbf{1.807} & \textbf{0.698} & 0.844 \\ \hline
\end{tabular}%
}
\end{table*}

\subsection{Qualitative analysis}
\label{Sec:Qualitative}
\begin{figure*}[t!]
    \centering
    \renewcommand{\arraystretch}{0.2} 
    \setlength{\tabcolsep}{0.2pt} 
    \begin{tabular}{cccccc}
        \multicolumn{1}{c}{{Input}} &
        \multicolumn{1}{c}{{Ground truth}} &
        \multicolumn{1}{c}{{SalNAS-XL(ours)}} &
        \multicolumn{1}{c}{{TranSalNet}} &
        \multicolumn{1}{c}{{EfficientNet-B4}} &
        \multicolumn{1}{c}{{TResNet-M}} \\

        \includegraphics[width=0.165\textwidth]{input_COCO_val2014_000000331712.jpg} &
        \includegraphics[width=0.165\textwidth]{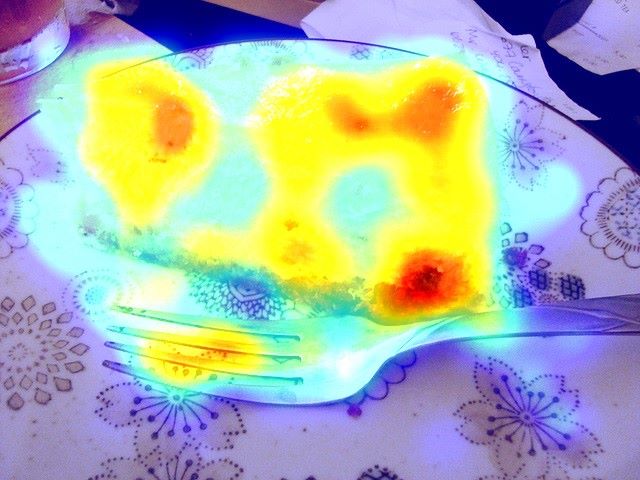} &
        \includegraphics[width=0.165\textwidth]{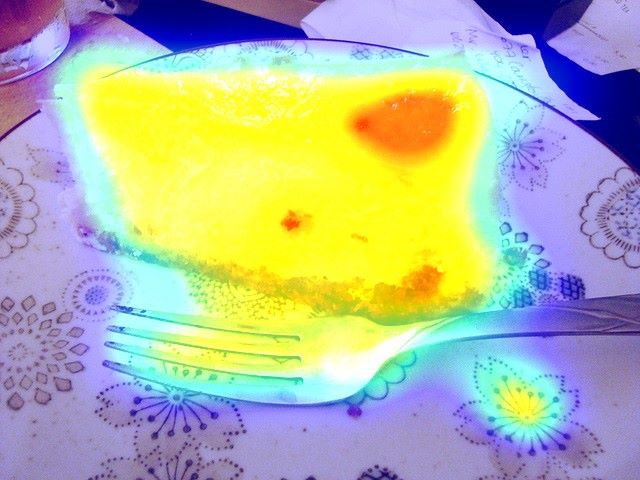} &
        \includegraphics[width=0.165\textwidth]{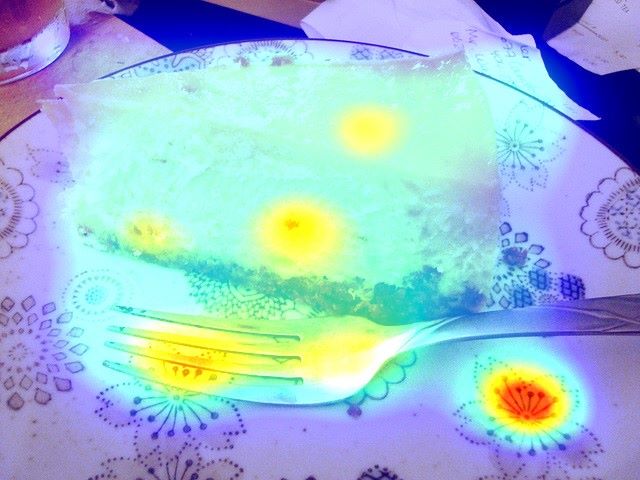} &
        \includegraphics[width=0.165\textwidth]{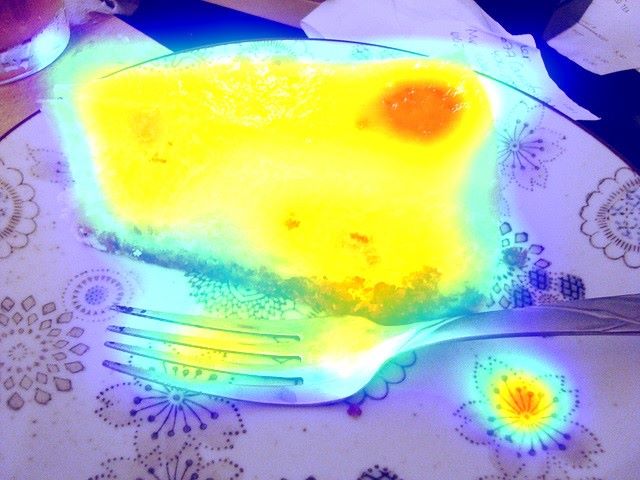} &
        \includegraphics[width=0.165\textwidth]{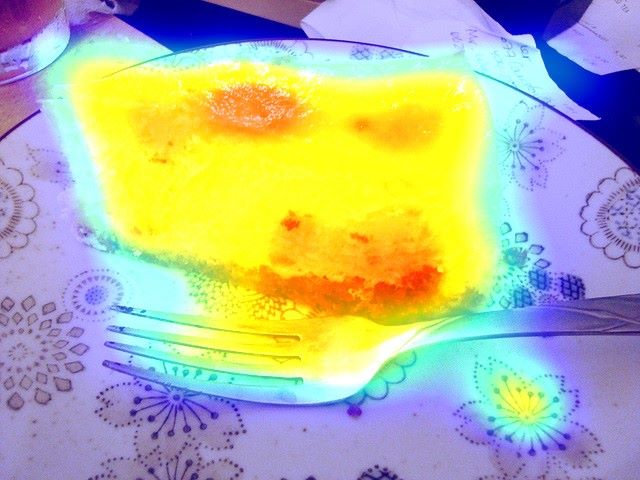} \\

	\\
	\multicolumn{1}{c}{\fontsize{8}{9}\selectfont{-}} &
	\multicolumn{1}{c}{\fontsize{8}{9}\selectfont{-}} &
	\multicolumn{1}{c}{\fontsize{8}{9}\selectfont{CC: 0.8962}} &
	\multicolumn{1}{c}{\fontsize{8}{9}\selectfont{CC: 0.6475}} &
	\multicolumn{1}{c}{\fontsize{8}{9}\selectfont{CC: 0.8838}} &
	\multicolumn{1}{c}{\fontsize{8}{9}\selectfont{CC: 0.8749}} \\
	\\

        \includegraphics[width=0.165\textwidth]{input_COCO_val2014_000000118065.jpg} &
        \includegraphics[width=0.165\textwidth]{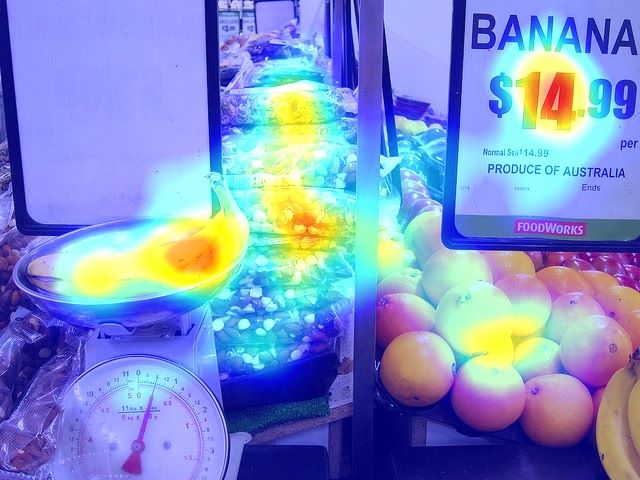} &
        \includegraphics[width=0.165\textwidth]{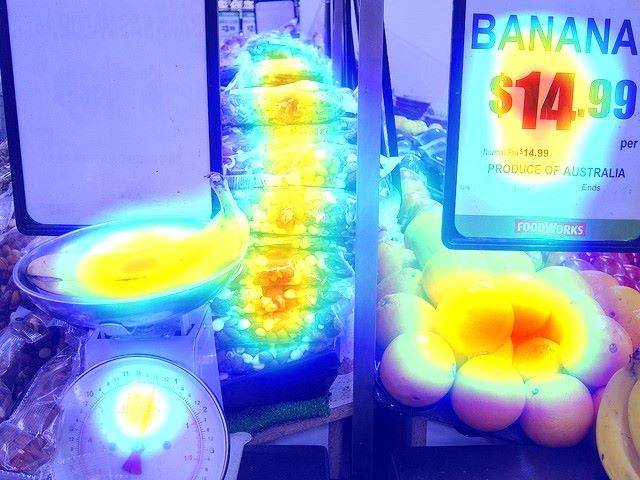} &
        \includegraphics[width=0.165\textwidth]{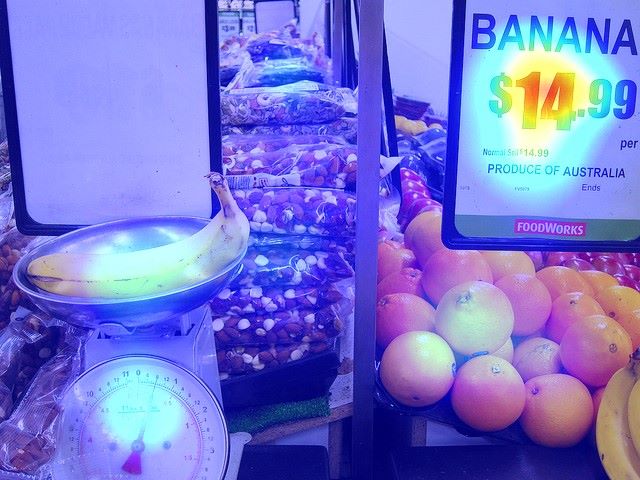} &
        \includegraphics[width=0.165\textwidth]{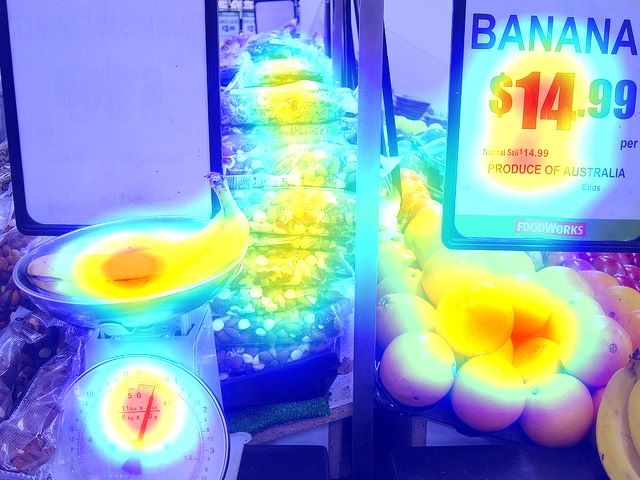} &
        \includegraphics[width=0.165\textwidth]{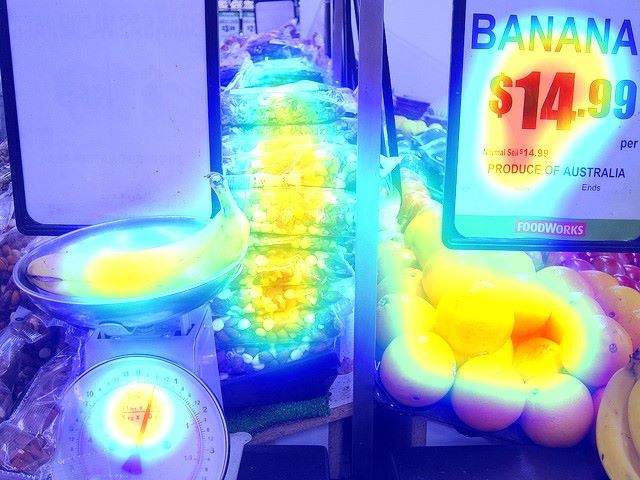} \\
        
	\\
	\multicolumn{1}{c}{\fontsize{8}{9}\selectfont{-}} &
	\multicolumn{1}{c}{\fontsize{8}{9}\selectfont{-}} &
	\multicolumn{1}{c}{\fontsize{8}{9}\selectfont{CC: 0.8471}} &
	\multicolumn{1}{c}{\fontsize{8}{9}\selectfont{CC: 0.6579}} &
	\multicolumn{1}{c}{\fontsize{8}{9}\selectfont{CC: 0.8026}} &
	\multicolumn{1}{c}{\fontsize{8}{9}\selectfont{CC: 0.7916}} \\
	\\

        \includegraphics[width=0.165\textwidth]{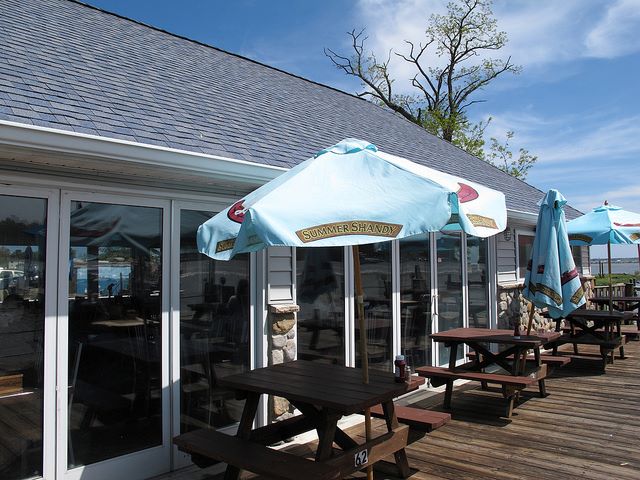} &
        \includegraphics[width=0.165\textwidth]{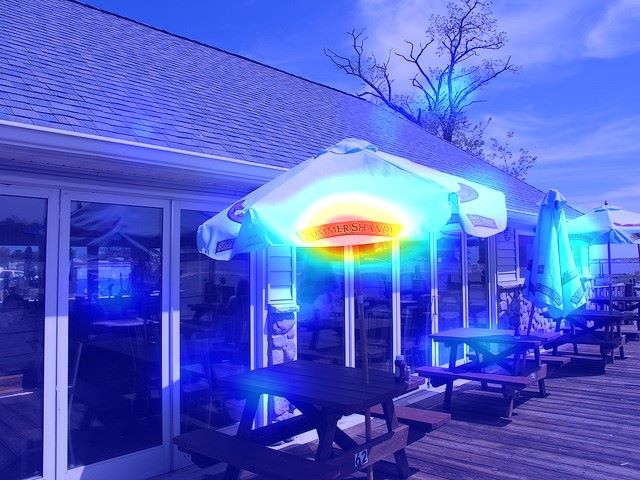} &
        \includegraphics[width=0.165\textwidth]{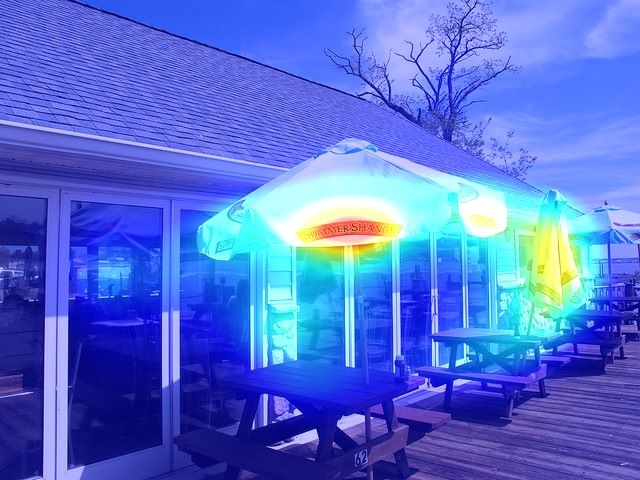} &
        \includegraphics[width=0.165\textwidth]{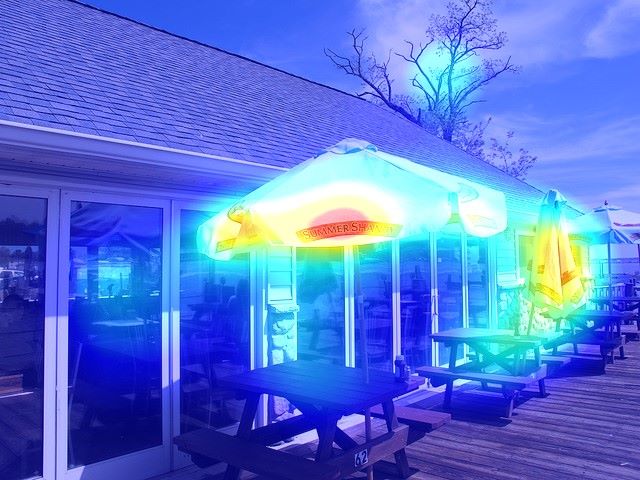} &
        \includegraphics[width=0.165\textwidth]{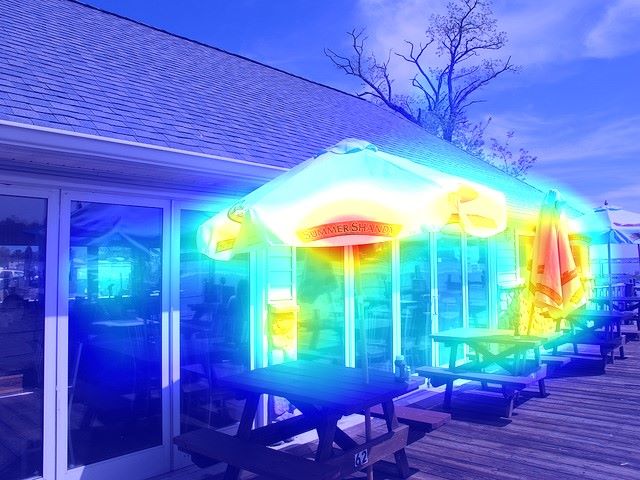} &
        \includegraphics[width=0.165\textwidth]{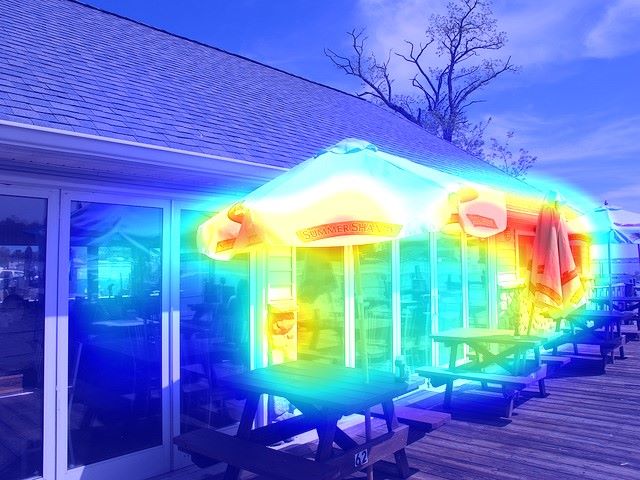} \\

	\\
	\multicolumn{1}{c}{\fontsize{8}{9}\selectfont{-}} &
	\multicolumn{1}{c}{\fontsize{8}{9}\selectfont{-}} &
	\multicolumn{1}{c}{\fontsize{8}{9}\selectfont{CC: 0.8299}} &
	\multicolumn{1}{c}{\fontsize{8}{9}\selectfont{CC: 0.8114}} &
	\multicolumn{1}{c}{\fontsize{8}{9}\selectfont{CC: 0.7879}} &
	\multicolumn{1}{c}{\fontsize{8}{9}\selectfont{CC: 0.6891}} \\
	\\

        \includegraphics[width=0.165\textwidth]{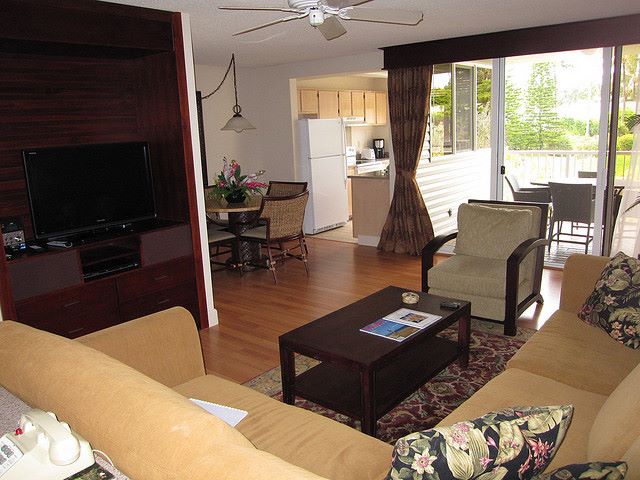} &
        \includegraphics[width=0.165\textwidth]{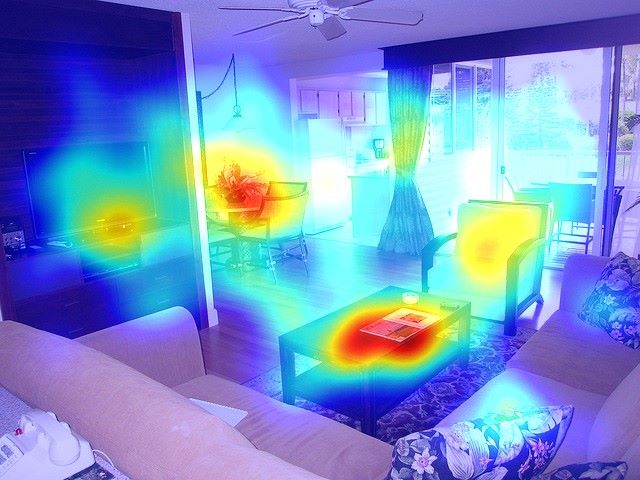} &
        \includegraphics[width=0.165\textwidth]{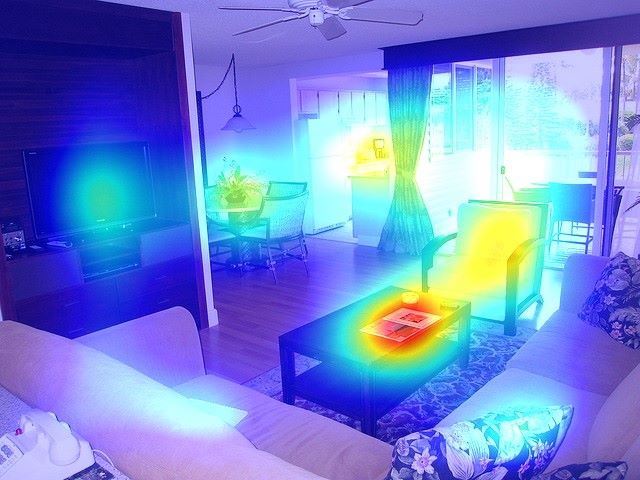} &
        \includegraphics[width=0.165\textwidth]{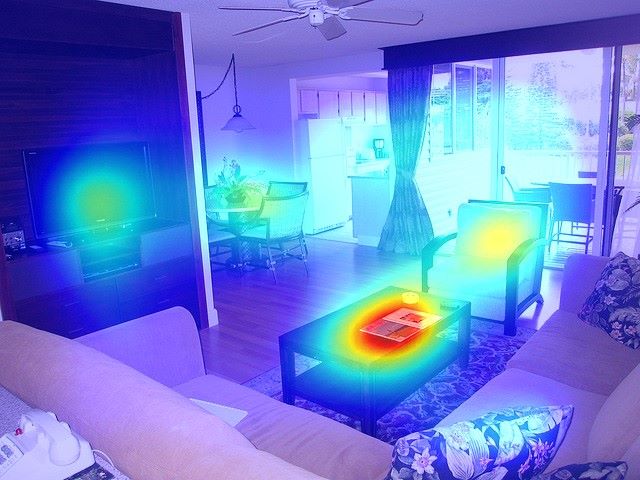} &
        \includegraphics[width=0.165\textwidth]{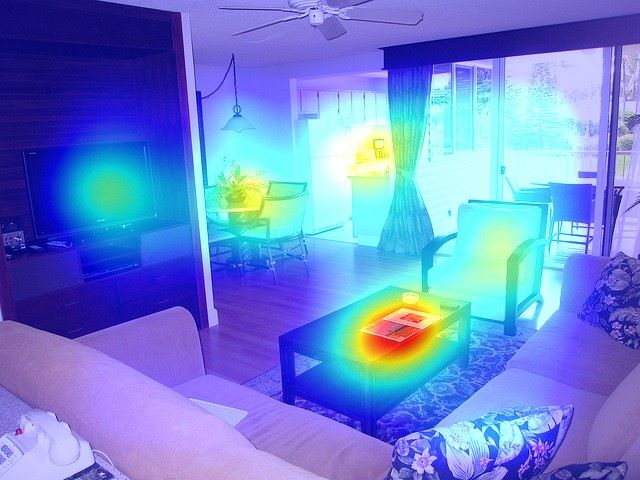} &
        \includegraphics[width=0.165\textwidth]{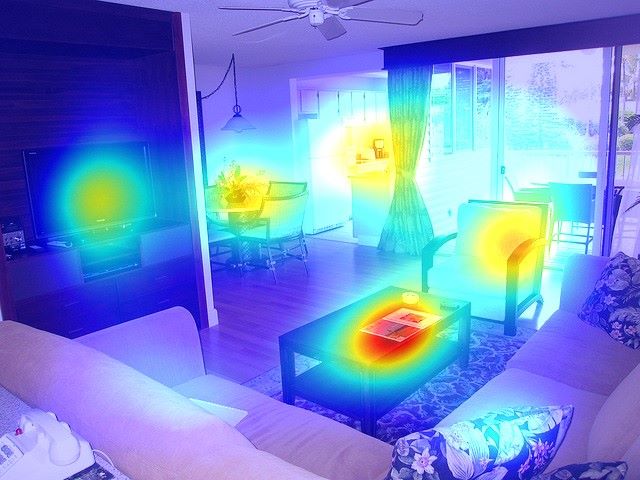} \\

	\\
	\multicolumn{1}{c}{\fontsize{8}{9}\selectfont{-}} &
	\multicolumn{1}{c}{\fontsize{8}{9}\selectfont{-}} &
	\multicolumn{1}{c}{\fontsize{8}{9}\selectfont{CC: 0.8163}} &
	\multicolumn{1}{c}{\fontsize{8}{9}\selectfont{CC: 0.8947}} &
	\multicolumn{1}{c}{\fontsize{8}{9}\selectfont{CC: 0.8834}} &
	\multicolumn{1}{c}{\fontsize{8}{9}\selectfont{CC: 0.8667}} \\
	\\

        \includegraphics[width=0.165\textwidth]{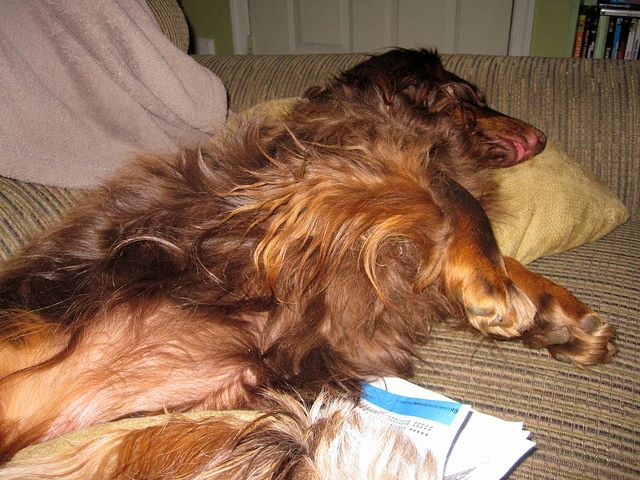} &
        \includegraphics[width=0.165\textwidth]{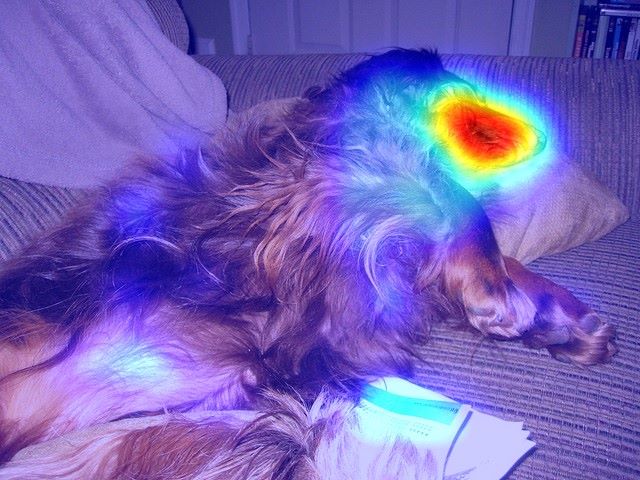} &
        \includegraphics[width=0.165\textwidth]{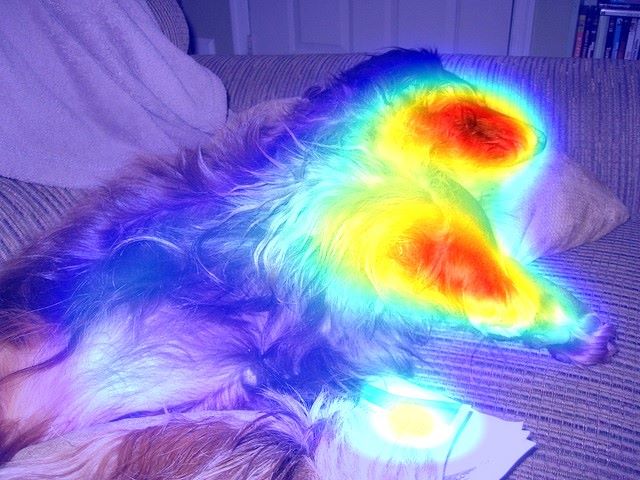} &
        \includegraphics[width=0.165\textwidth]{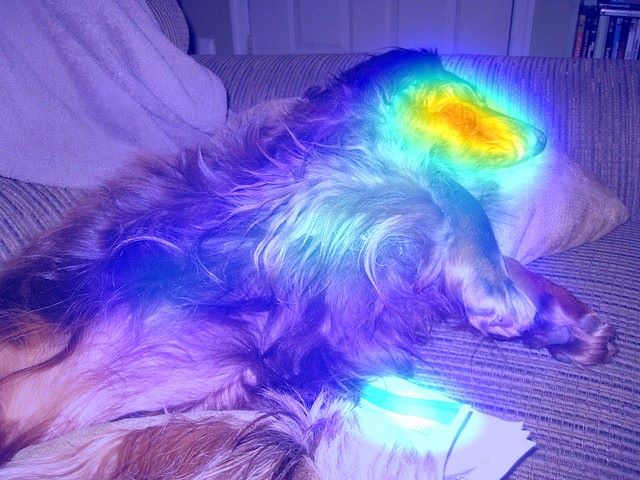} &
        \includegraphics[width=0.165\textwidth]{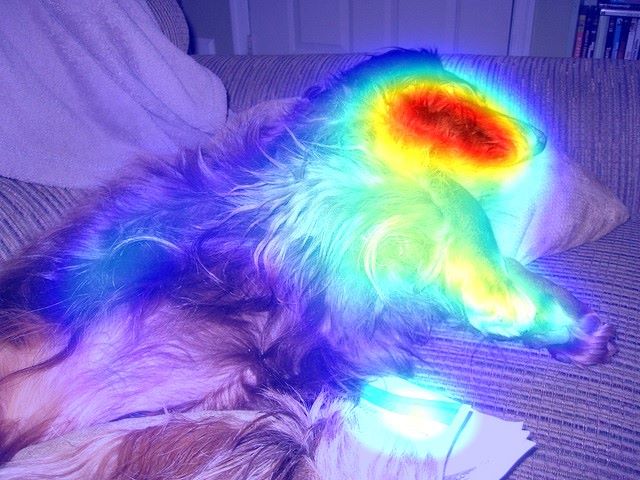} &
        \includegraphics[width=0.165\textwidth]{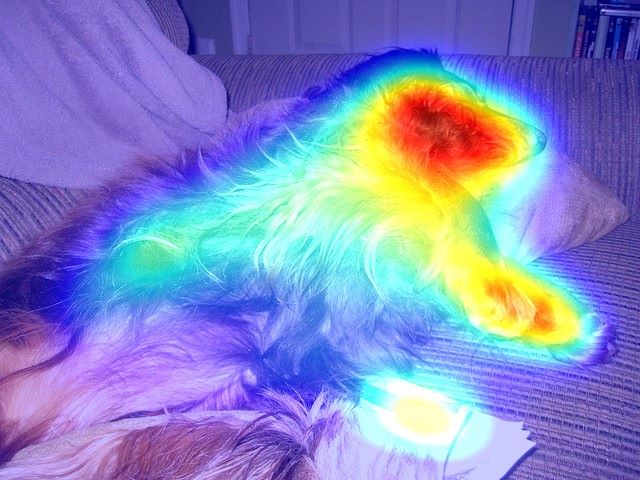} \\

	\\
	\multicolumn{1}{c}{\fontsize{8}{9}\selectfont{-}} &
	\multicolumn{1}{c}{\fontsize{8}{9}\selectfont{-}} &
	\multicolumn{1}{c}{\fontsize{8}{9}\selectfont{CC: 0.7090}} &
	\multicolumn{1}{c}{\fontsize{8}{9}\selectfont{CC: 0.8501}} &
	\multicolumn{1}{c}{\fontsize{8}{9}\selectfont{CC: 0.8073}} &
	\multicolumn{1}{c}{\fontsize{8}{9}\selectfont{CC: 0.6856}} \\
	\\

        \includegraphics[width=0.165\textwidth]{input_COCO_val2014_000000523195.jpg} &
        \includegraphics[width=0.165\textwidth]{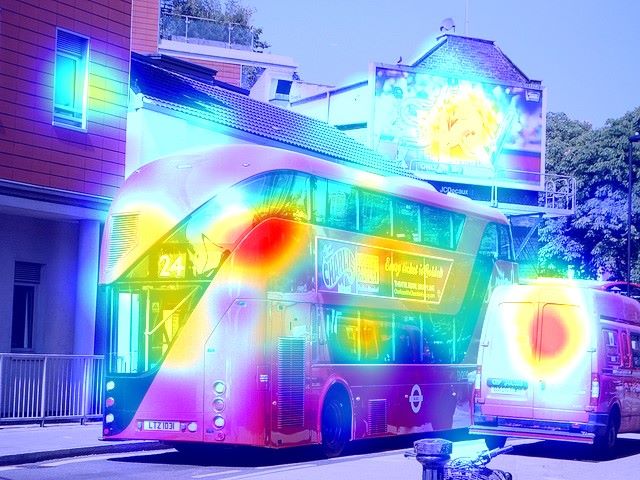} &
        \includegraphics[width=0.165\textwidth]{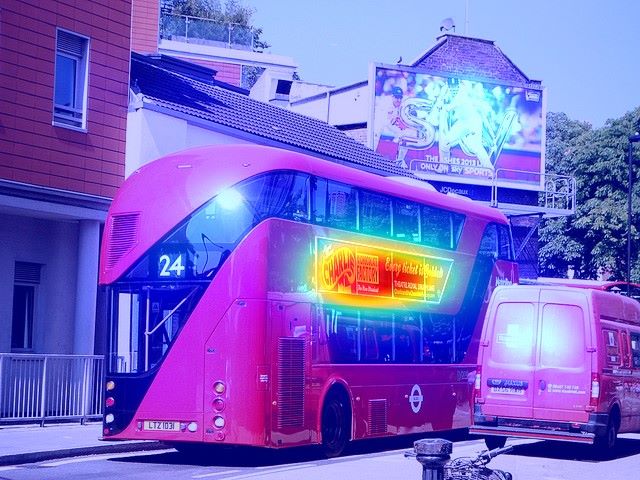} &
        \includegraphics[width=0.165\textwidth]{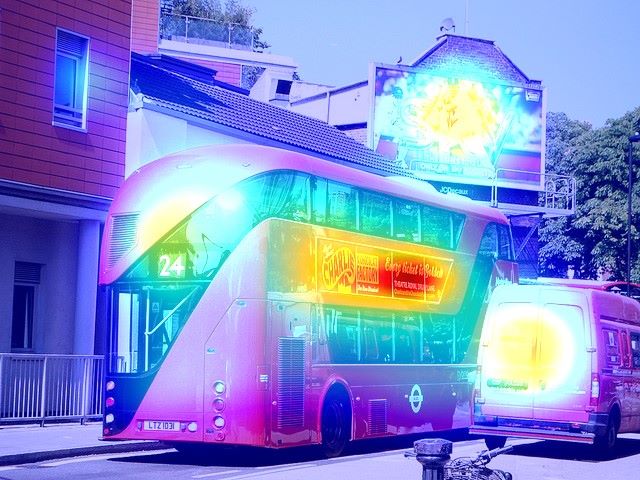} &
        \includegraphics[width=0.165\textwidth]{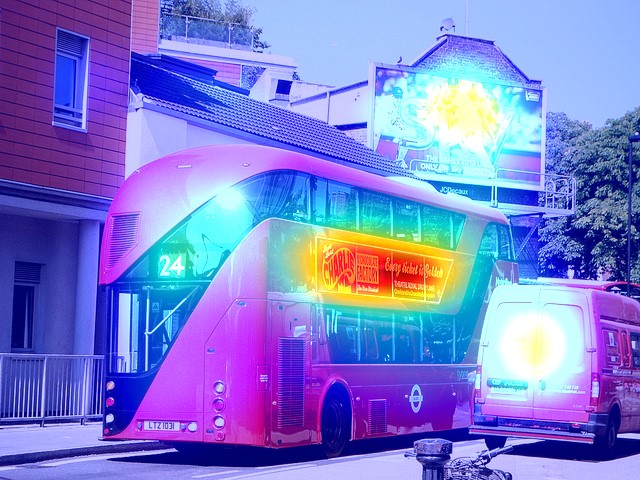} &
        \includegraphics[width=0.165\textwidth]{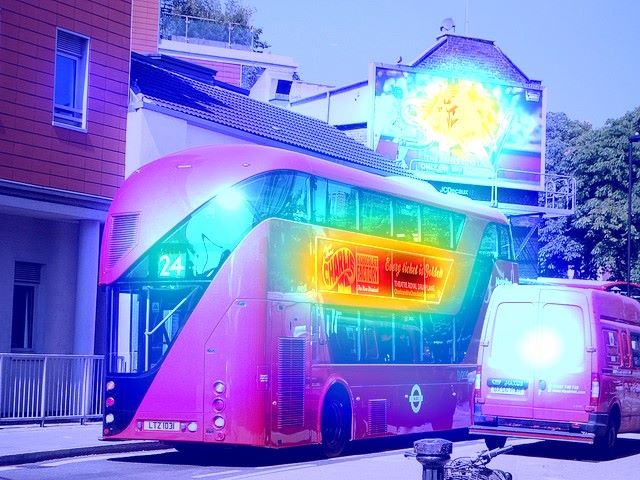} \\
        
	\\
	\multicolumn{1}{c}{\fontsize{8}{9}\selectfont{-}} &
	\multicolumn{1}{c}{\fontsize{8}{9}\selectfont{-}} &
	\multicolumn{1}{c}{\fontsize{8}{9}\selectfont{CC: 0.6153}} &
	\multicolumn{1}{c}{\fontsize{8}{9}\selectfont{CC: 0.8545}} &
	\multicolumn{1}{c}{\fontsize{8}{9}\selectfont{CC: 0.7874}} &
	\multicolumn{1}{c}{\fontsize{8}{9}\selectfont{CC: 0.7854}} \\

    \end{tabular}
    \caption{Qualitative comparison between our model (SalNAS-XL) and other models, including TranSalNet, EfficientNet-B4, and TResNet-M. Images are sourced from the SALICON validation dataset. Rows 1-3 showcase the saliency maps produced by the SalNAS-XL subnet, closely matching ground truth. Rows 4-6 of SalNAS-XL manifest lower \acrshort{cc} scores compared to other models.}
    \label{fig:3}
\end{figure*}

We employ the SALICON validation dataset to provide the visualization results, shown in Fig.~\ref{fig:3}. We compared SalNAS-XL with state-of-the-art saliency prediction models, \ie, TranSalNet, TResNet-M, and state-of-the-art backbone, EfficientNet-B4. For each prediction, a \acrshort{cc} score is provided, indicating the performance of each model. A higher \acrshort{cc} score shows better alignment of ground truth with the prediction. Our SalNAS-XL provides the precise area of attention with similar magnitudes to the ground truth. \\

We have chosen two sets of sample images. The first set comprises images that the SalNAS-XL model can predict accurately. The second set consists of images for which the SalNAS-XL model's predictions are less accurate than others, as follows: rows 1 to 3 demonstrate the performance of SalNAS-XL against other state-of-the-art. The results show that SalNAS-XL provides saliency maps that closely align with the ground truth; the \acrshort{cc} scores further support this observation, indicating that higher \acrshort{cc} scores correspond to better prediction. Furthermore, for rows 4 to 6, SalNAS-XL provides a lower \acrshort{cc} score compared to other models, indicating a weaker alignment between the predicted saliency map and the ground truth.

In addition, a closer look at row 5 shows that the ground truth depicts a single attention peak at the dog's face, while the SalNAS-XL saliency map represents additional peaks. Other models show better alignment with the ground truth. Similarly, in the sample from row 6, the ground truth displays multiple peaks, while SalNAS-XL fails to produce multiple peaks compared to the other models.


\begin{table*}[t!]
\caption{Real-time processing evaluation of various saliency prediction models and state-of-the-art backbones.}
\label{tab:realtime}
\begin{threeparttable}
\resizebox{\textwidth}{!}{%
\begin{tabular}{lcccccccc}
\hline
\multicolumn{1}{c}{\textbf{Model}} & \textbf{Input Size} & \textbf{\begin{tabular}[c]{@{}c@{}}FLOPS\\ (G) $\downarrow$\end{tabular}} & \textbf{\begin{tabular}[c]{@{}c@{}}Params\\ (M) $\downarrow$\end{tabular}} & \textbf{\begin{tabular}[c]{@{}c@{}}Size\\ (MB) $\downarrow$\end{tabular}} & \textbf{\begin{tabular}[c]{@{}c@{}}Emission\\ $\mathbf{(gCO_{2}eq)}$ $\downarrow$\end{tabular}} & \textbf{\begin{tabular}[c]{@{}c@{}}Power consumption\\ (W) $\downarrow$\end{tabular}} & \textbf{\begin{tabular}[c]{@{}c@{}}Latency\\ M2 (ms/image) $\downarrow$\end{tabular}} & \textbf{\begin{tabular}[c]{@{}c@{}}Throughput\\ RTX (images/s) $\uparrow$\end{tabular}} \\ \hline
EEEA-Net-C2 & 256 × 192 & 0.59 & \textbf{4.57} & \textbf{17.54} & 0.26 & 0.44 & 28.57 & \textbf{3042} \\
OFA595 & 256 × 192 & 1.1 & 7.41 & 28.45 & 0.38 & 0.66 & 35.18 & 2053 \\
EfficientNet-B0 & 256 × 192 & 1.23 & 5.44 & 20.91 & 0.38 & 0.65 & 26.21 & 2074 \\
SalNAS-XS (ours) & 256 × 192 & \textbf{0.56} & 6.35 & 24.34 & \textbf{0.24} & \textbf{0.41} & \textbf{25.27} & 2930 \\ \hline
SimpleNet & 256 × 256 & 35.72 & 116.21 & 443.98 & 2.91 & 4.99 & 129.63 & 283 \\
SalFBNet & 384 × 224 & 76.29 & 17.78 & 67.84 & 4.5 & 7.72 & 220.32 & 178 \\ \hline
TranSalNet & 384 × 288 & 47.64 & 76.56 & 292.89 & 5.53 & 9.49 & 152.51 & 171 \\
EfficientNet-B4 & 384 × 288 & 7.50 & 20.41 & 78.35 & 0.76 & 1.50 & 63.90 & 474 \\
EfficientNet-B7 & 384 × 288 & 18.89 & 69.51 & 266.34 & 4.59 & 7.87 & 75.92 & 478 \\
SalNAS-XL (ours) & 384 × 288 & \textbf{3.88} & \textbf{20.98} & \textbf{80.42} & \textbf{1.95} & \textbf{3.35} & \textbf{43.53
} & \textbf{497} \\ \hline
\end{tabular}
}
\begin{tablenotes}
\small {
\item M2 represents Apple M2 system on a chip (SoC); RTX represents NVIDIA RTX 4090 GPU. 
}
\end{tablenotes}
\end{threeparttable}
\end{table*}

\begin{figure}[t!]
\begin{center}
\includegraphics[width=0.97\linewidth]{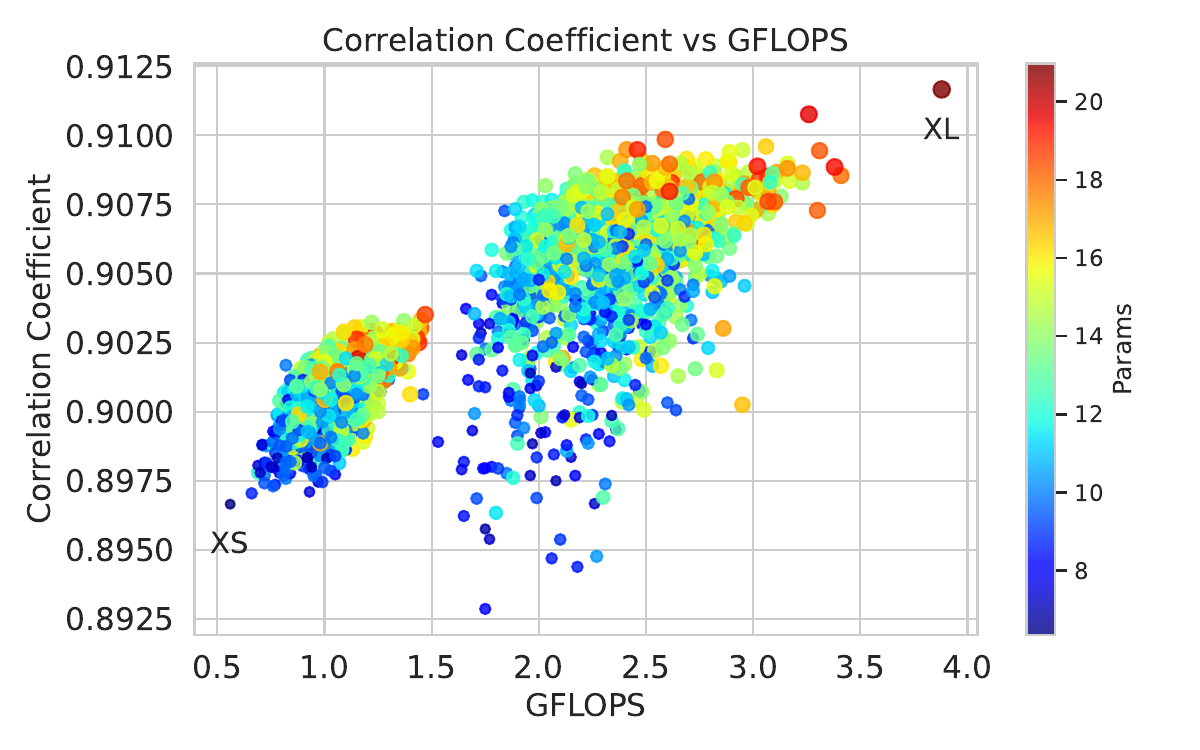}
\end{center}
\caption{Exploring the search space of SalNAS: a bubble chart comparing computational complexity (GFLOPS) and correlation coefficient for a sample of 2000 subnets. Bubble size indicates parameter count, and the chart displays various model designs, ranging from SalNAS-XS to SalNAS-XL.}
\label{fig:searchspace}
\end{figure}

\subsection{Real-time processing performance}
\label{Sec:Realtime}
Table~\ref{tab:realtime} illustrates the real-time processing performance of various state-of-the-art saliency prediction models. All results are obtained using the following hardware specifications 13th Gen Intel(R) Core (TM) i9-13900KF CPU, DDR5 64 GB RAM, and NVIDIA GeForce RTX 4090 GPU. Real-time processing performance is evaluated using seven assessment measures: computational complexity, parameters, model size, carbon emission, power consumption, latency, and throughput. We considered running each image for latency calculation 200 times on an Apple M2 system on a chip (SoC), where each batch size is one and takes the average latency value. To calculate the carbon footprint of our computing infrastructure in Thailand (THA), we monitored the energy utilization of the CPU, GPU, and RAM during our experiments by forwarding 200 times a batch of size 64 to the model. To accurately assess its environmental effect, the resulting data is analyzed using CodeCarbon, a Python package that enables real-time power monitoring and estimates carbon emissions associated with software applications.

We evaluated two subnets, namely SalNAS-XS (small-size) and SalNAS-XL (large-size), derived from our proposed supernet SalNAS, as illustrated in Fig.~\ref{fig:searchspace}, in comparison with state-of-the-art methods. Our SalNAS-XS is compared against three state-of-the-art backbones---EEEA-Net-C2, OFA595, and EfficicnetNet-B0---with input image resolution of 256$\times$192. Our SalNAS-XL is compared against the state-of-the-art backbones---EfficicnetNet-B4 and EfficicnetNet-B7---and state-of-the-art saliency prediction, TranSalNet, with the input resolution of 384$\times$288. Table~\ref{tab:realtime} shows that SalNAS-XS offers competitive computational costs in computational complexity, carbon emission, and power consumption. Meanwhile, SalNAS-XL outperforms all the state-of-the-art saliency prediction backbone, by a significant margin in the number of FLOPS, latency, power consumption, carbon emission, parameters, model size, and throughput.

\subsection{Limitation}
The proposed work has two limitations. Firstly, high computational cost associated with model search across sub-networks. Search strategies such as reinforcement learning and evolutionary algorithms require high computational costs, which consequently affect time and energy efficiency. These algorithms can be improved by optimizing the neural architecture generators to dynamically produce diverse components while keeping the dimensionality of the problem low. This can result in a wider and more cost-effective search. These generators can assist in developing optimal models customized for specific deployment scenarios, consequently improving search efficiency and flexibility. Secondly, Self-KD lacks external supervision, making it difficult to handle complex tasks without a teacher model. It may not be suitable for all models or applications, particularly where high accuracy is required. Traditional KD is recommended in such cases for better student model performance. However, this paper's SalNAS model overcomes this limitation by using weight-sharing, treating the SalNAS-XL model as a teacher for smaller subnets. Thus, Self-KD with NAS is recommended for performance gains.

\section{Conclusion}
We proposed a novel supernet SalNAS, a weight-sharing network that includes all candidate architectures by incorporating a dynamic convolution into the encoder-decoder design for saliency prediction. Our proposed SalNAS-XL is the biggest subnet of our supernet, yet with high efficiency in the number of parameters and computational complexity. Nevertheless, due to the limited number of parameters, our subnet can still struggle to generalize effectively across diverse datasets. Therefore, we propose Self-KD to regularize the student knowledge with the best-performing parameters chosen via cross-validation. This results in the highest performance across seven saliency prediction benchmark datasets---SALICON, CAT2000, MIT1003, DUT-OMRON, PASCAL-S, FIWI, and OSIE---where our performance comparison is made against ten saliency prediction and five state-of-the-art backbones, \eg, EEEA-Net-C2, OFA595, MobileNetV3, EfficientNet-B4, and TResNet-M. We also provide in-depth studies which reveal that our Self-KD strategy not only improves the generalization performance of SalNAS but also the performance of other classification and saliency prediction backbones. Finally, we evaluated SalNAS for real-time processing by considering seven real-time assessment measures, \ie, computational complexity, parameters, model size, carbon emission, power consumption, latency, and throughput. Our result indicates that SalNAS-XL outperforms all state-of-the-art models across all assessment measures. 

A hybrid supernet between the CNN and Vision transformer architectures with federated learning can be considered in the future. This advancement can help to improve real-world applications such as medical diagnostics, smart surveillance, and video enhancement. In addition, the lack of large-scale annotated datasets can be mitigated by using semi-supervised learning to leverage both labeled and unlabeled data for better model performance.

\section*{Acknowledgements}
The authors would like to acknowledge the Division of Research and Innovation at Naresuan University for their support via Grant No. R2566C011, as well as the Global and Frontier Research University Fund at Naresuan University under Grant No. R2567C002, which partially supported this research project.

\bibliography{mybibfile}



\end{document}